\documentclass[a4paper]{article}

\usepackage[english]{babel}
\usepackage[utf8x]{inputenc}
\usepackage[T1]{fontenc}
\usepackage{enumitem}
\usepackage{amsthm}
\usepackage{amsmath}
\usepackage{amsfonts} 
\usepackage{algorithm, algorithmicx, algpseudocode}
\usepackage{graphicx} 
\usepackage{multirow,array}
\usepackage{makecell}
\usepackage{pinyin}
\usepackage[normalem]{ulem}
\usepackage[encapsulated]{CJK}
\usepackage{mathtools}
\usepackage{graphicx,color,xcolor}
\usepackage{enumitem} 
\usepackage{bm}
\usepackage{algorithm}
\usepackage{tablefootnote}
\usepackage{relsize}
\usepackage{footnote}
\usepackage{bbding}
\usepackage{xcolor,colortbl}
\makesavenoteenv{tabular}
\usepackage[encapsulated]{CJK}
\usepackage{authblk}
\usepackage{placeins}
\usepackage{comment}
\usepackage{booktabs}
\usepackage{subcaption}
\usepackage{url}

\usepackage[a4paper,top=3cm,bottom=2cm,left=3cm,right=3cm,marginparwidth=1.75cm]{geometry}

\usepackage{amsmath}
\usepackage{graphicx}
\usepackage[colorinlistoftodos]{todonotes}
\usepackage[colorlinks=true, allcolors=blue]{hyperref}

	\title{Achieving Human Parity on Automatic\\ Chinese to English News Translation}
\author {Hany Hassan\footnote{Corresponding author: hanyh@microsoft.com}}
\author {Anthony Aue}
\author {Chang Chen}
\author {Vishal Chowdhary}
\author{Jonathan Clark}
\author{Christian Federmann}
\author{Xuedong Huang}
\author{Marcin Junczys-Dowmunt}
\author{William Lewis}
\author{Mu Li}
\author{Shujie Liu}
\author{Tie-Yan Liu}
\author{Renqian Luo}
\author{Arul Menezes}
\author{Tao Qin}
\author{Frank~Seide}
\author{Xu Tan}
\author{Fei Tian}
\author{Lijun Wu}
\author{Shuangzhi Wu}
\author{Yingce Xia}
\author{Dongdong~Zhang}
\author{Zhirui Zhang}
\author{Ming Zhou}

\affil{Microsoft AI \& Research} 
\date{}

\begin{document}
\begin{CJK*}{UTF8}{gbsn}
\maketitle

\begin{abstract}
Machine translation has made rapid advances in recent years. Millions of people are using it today in online translation systems and mobile applications in order to communicate across language barriers. The question naturally arises whether such systems can approach or achieve parity with human translations. In this paper, we first address the problem of how to define and accurately measure human parity in translation. We then describe Microsoft's machine translation system and measure the quality of its translations on the widely used WMT~2017 news translation task from Chinese to English. We find that our latest neural machine translation system has reached a new state-of-the-art, and that the translation quality is at human parity when compared to professional human translations. We also find that it significantly exceeds the quality of crowd-sourced non-professional translations.
\end{abstract}
\section{Introduction}
\label{sec.into}
Recent years have seen human performance levels reached or surpassed in tasks ranging from games such as Go \cite{Silver2016mastering} to classification of images in ImageNet \cite{DeepResidual} to conversational speech recognition on the Switchboard task \cite{xiong2017toward}. 

In the area of machine translation, we have seen dramatic improvements in quality with the advent of attentional encoder-decoder neural networks \cite{sutskever2014sequence,bahdanau2014neural,vaswani2017attention}. However, translation quality continues to vary a great deal across language pairs, domains, and genres, more or less in direct relationship to the availability of training data. This paper summarizes how we achieved human parity in translating text in the news domain, from Chinese to English. While the techniques we used are not specific to the news domain or the Chinese-English language pair, we do not claim that this result necessarily generalizes to other language pairs and domains, especially where limited by the availability of data and resources.

Translation of news text has been an area of active interest in the Machine Translation community for over a decade, due to the practical and commercial importance of this domain, the availability of abundant parallel data on the web (at least in the most popular languages) and a long history of government-funded projects and evaluation campaigns, such as NIST-OpenMT\footnote{https://www.nist.gov/itl/iad/mig/open-machine-translation-evaluation} and GALE\footnote{https://www.nist.gov/itl/iad/mig/machine-translation-evaluation-gale}. The annual evaluation campaign of the WMT (Conference on Machine Translation) \cite{bojar-EtAl:2017:WMT1}, has also focused on news translation for more than a decade.

Defining and measuring human quality in translation is challenging for a number of reasons. Traditional metrics of translation quality, such as BLEU \cite{papineni2002bleu}, TER \cite{snover2006study} and Meteor \cite{denkowskimeteor2011} measure translation quality by comparison with one or more human reference translations. However, the same source sentence can be translated in sometimes substantially different but equally correct ways.  This makes reference-based evaluation nearly useless in determining quality of human translations or near-human-quality machine translations. 

Further complicating matters, we find that the quality of reference translations, long assumed to be "gold" annotations by professional translators, are sometimes of remarkably poor quality. This is because references are often crowd-sourced (either directly, or indirectly through translation vendors). We have observed that crowd workers often use on-line MT with or without post-editing, rather than translating from scratch. Furthermore, many crowd workers appear to have only a rudimentary grasp of one of the languages, which often leads to unacceptable translation quality.

In Section \ref{sec:human-parity}, we describe how we address these challenges in defining and measuring human quality. In Section \ref{sec:sys-desc}, we describe our system architecture. Section \ref{sec:experiments} describes our data and experiments. Sections \ref{sec:human-eval-results} and \ref{human-analysis} present our evaluation results and analysis.  

\section{Human Parity on Translation}
\label{sec:human-parity}

\newtheorem{definition}{Definition}

Achieving human parity for machine translation is an important milestone of machine translation research. However, the idea of computers achieving human quality level is generally considered unattainable and triggers negative reactions from the research community and end users alike. This is understandable, as previous similar announcements have turned out to be overly optimistic.

Before any meaningful discussion of human parity can occur, we require a rigorous definition of the concept of human parity for translation. Based on this theoretical definition we can then investigate how close neural machine translation is to this goal.


\subsection{Defining Human Parity}

Intuitively, we can define human parity for translation as follows:

\begin{definition}
If a bilingual human judges the quality of a candidate translation produced by a human to be equivalent to one produced by a machine, then the machine has achieved \textsc{human parity}.
\end{definition}






Assuming that it is possible for humans to measure translation quality by assigning scores to translations of individual sentences of a test set, and generalizing from a single sentence to a set of test sentences, this effectively yields the following statistical definition:

\begin{definition}
\label{def:human-parity}
If there is no statistically significant difference between human quality scores for a test set of candidate translations from a machine translation system and the scores for the corresponding human translations then the machine has achieved \textsc{human parity}.
\end{definition}


We choose definition~\ref{def:human-parity} to address the question of human parity for machine translation in a fair and principled way. Given a reliable scoring metric to determine translation quality, based on direct human assessment, one can use a paired statistical significance test to decide whether a given machine translation system can be considered at parity with human translation quality for a test set and corresponding human references.

It is important to note that this definition of human parity does not imply that the machine translation system \emph{outperforms} the human benchmark, but rather that its quality is statistically \emph{indistinguishable}. It also does not imply that the translation is error-free. Machines, like humans, will continue to make mistakes.

Finally, achieving human parity on a given test set is measured with respect to a specific set of benchmark human translations and does not automatically generalize to other domains or language pairs. 

\subsection{Judging Human Parity}
Our operational definition of human parity requires that human annotators be used to judge translation quality. While there exist various automated metrics to measure machine translation quality, these can only act as a (not necessarily correlated) proxy. Such metrics are typically reference-based and thus subject to \emph{reference bias}. This can occur in the form of bad reference translations which result in bad segment scores. Also, due to the generative nature of translation, there often are multiple valid translations for a given input segment. Any translation which does not closely match the structure of the corresponding reference has a scoring disadvantage, even perfect human translations. While these effects can be lessened using multiple references, the underlying problem remains unsolved\footnote{HyTER~\cite{dreyer2012hyter} attempted to solve this but did not achieve mainstream success.}.




Therefore, following the Conference on Machine Translation (WMT17)~\cite{bojar-EtAl:2017:WMT1}, we adopt \emph{direct assessment}~\cite{yvetteDA}  as our human evaluation method. To avoid reference bias---which can also happen for human evaluation\footnote{Results from both source-based and reference-based direct assessment collected for IWSLT17~\cite{IWSLT17} show that annotators assign higher scores in the source-based scenario and that they are more strict with their scoring in the reference-based scenario. This indicates that references do in fact influence human scoring behavior. Consequently, bad references will affect human evaluation in a reference-based direct assessment.}---we use the \emph{source-based} evaluation methodology following IWSLT17~\cite{IWSLT17}. 

In source-based \emph{direct assessment}, annotators are shown source text and a candidate translation and are asked the question \emph{``How accurately does the above candidate text convey the semantics of the source text?''}, answering this using a slider ranging from 0 (\emph{Not at all}) to 100 (\emph{Perfectly}).\footnote{Co-author Christian Federmann, in his role as co-organizer of the annual WMT evaluation campaign, was instrumental in developing the Appraise evaluation system used by WMT and also in this paper. He was not involved in developing the systems being evaluated here, nor were the human benchmark references available to the system developers. Hence, our evaluation was implemented in a double-blind manner.} As a side effect, we have to employ bilingual annotators for our human evaluation campaigns.
%
%
%
%

The raw human scores are then standardized to a $z$-score, defined as the signed number of standard deviations an observation is above the mean, relative to a sample.

The $z$-scores are then averaged at the segment and system level. Results with statistically insignificant differences are grouped into clusters (according to Wilcoxon rank sum test \cite{wilcoxon1945individual} at p-level $p \leq 0.05$).\footnote{WMT17 implemented this using R's \texttt{wilcox.test()}. Our implementation differs from this as the clustering has been integrated into Appraise and uses the Mann-Whitney rank test \cite{mann1947test} at the same p-level $p \leq 0.05$, based on Python's \texttt{scipy.mannwhitneyu()}. For the purpose of determining if the difference between scores for two candidate systems is statistically significant, both implementations are equivalent. \label{da-implementation-diffs}}


To identify unreliable crowd workers, direct assessment includes artificially degraded translation output, so called ``bad references''. Any large scale crowd annotation task requires such integrated quality controls to guarantee high quality results. In our evaluation campaigns for Chinese into English, we observed relatively few attempts of gaming or spamming compared to other languages for which we run similar annotation tasks (we do not report on those in the context of this paper). In the remainder of this paper, direct assessment ranking clusters are computed in the same way as they had been generated for the WMT17 conference, with minor modifications\textsuperscript{\ref{da-implementation-diffs}}.

\section{System Description}
\label{sec:sys-desc}
\subsection{Neural Machine Translation}
Neural Machine Translation (NMT)~\cite{bahdanau2014neural} represents the state-of-the-art for translation quality. This has been demonstrated in various research evaluation campaigns (e.g.~WMT \cite{bojar-EtAl:2017:WMT1}), and also for large scale production systems \cite{wu2016google,devlin:2017:EMNLP2017}. NMT scales to train on parallel data on the order of tens of millions of sentences.

Currently, State-of-the-art NMT~\cite{bahdanau2014neural,sutskever2014sequence} is generally based on a sequence-to-sequence encoder-decoder model  with an attention mechanism \cite{bahdanau2014neural}. Attentional sequence-to-sequence NMT models the conditional probability $p(\mathbf{y}|\mathbf{x})$ of the translated sequence $\mathbf{y}$ given an input sequence $\mathbf{x}$.  
In general, an attentional NMT system $\theta$ consists of two components: an encoder $\theta_e$ which transforms the input sequence into a sequence or set of continuous representations,
and a decoder $\theta_d$ that dynamically reads out the encoder's output with an attention mechanism and predicts the conditional distribution of each target word. 
Generally, $\theta$ is trained to maximize the likelihood on a parallel training set consisting of $N$ sentence pairs: 
\begin{eqnarray}
	\mathcal{L}(\theta) & = &
    	\sum_{n=1}^N\log p\left(\mathbf{y}^{(n)}\bigl|\bigr.
        					\mathbf{x}^{(n)}; \theta\right) \nonumber\\
    & = &
    	\sum_{n=1}^N\sum_{t=1}^T\log p\left(y_t^{(n)}\bigl|\bigr.
        	\mathbf{y}_{<t}^{(n)}, h_{t-1}^{(n)},
        	f^{\mathrm{att}}\left(
            	f^{\mathrm{enc}}\bigl(\mathbf{x}^{(n)}\bigr),
                	\mathbf{y}_{<t}^{(n)}, h_{t-1}^{(n)},
            \right);
            \theta\right)
	 \label{eq.loss} 
\end{eqnarray}
where $h_{t-1}^{(n)}$ denotes an internal decoder state, and $\mathbf{y}_{<t}$ the words preceding step $t$. At each step $t$, the attention mechanism $f^{\mathrm{att}}$ determines a context vector as a weighted sum over the outputs of the encoder $f^{\mathrm{enc}}\bigl(\mathbf{x}^{(n)}\bigr)$, where the weights
are determined essentially by comparing each of the encoder's outputs
against the decoder's internal state and output up to time $t-1$.
$f^{\mathrm{enc}}$ is a sentence-level feature extractor and can be implemented as multi-layer bidirectional RNNs \cite{bahdanau2014neural,wu2016google}, a convolutional model (ConvS2S), \cite{gehring2017convolutional} or a Transformer \cite{vaswani2017attention}. 

Like RNN sequence-to-sequence models, ConvS2S and Transformer utilize an encoder-decoder architecture. However, both models aim to eliminate the internal decoder state $h_{t-1}$. This side steps the recurrent nature of RNN, in which each sentence is encoded word by word, which limits the parallelizability of the computation and makes the encoded representation  sensitive to the sequence length. 

ConvS2S utilizes a stacked convolutional representation that models the dependencies between nearby words on lower layers, while longer-range dependencies are handled in the upper layers of the stack. The decoder applies attention on each layers. ConvS2S also utilizes position sensitive embeddings along with residual connections to accommodate positional variance.

The Transformer model replaces the convolutions with self-attention, which also eliminates the recurrent processing and positional dependency in the encoder. It also utilizes multi-head attention, which allows to attend to multiple source positions at once, in order to model different types of dependencies regardless of position. Similar to ConvS2S, the Transformer model utilizes positional embeddings to compensate for the ordering information, though it proposes a non-parametric representation. While these models eliminate recurrence in the encoder, all models discussed above decode auto-regressively, where each output word's distribution is conditioned on previously generated outputs. The Transformer model has shown \cite{vaswani2017attention} to yield significant improvement and therefore was choses as the base for our work in this paper.

\subsection{Reaching Human Parity}

Despite immense progress on NMT in the research community over the past years, human parity has remained out of reach. In this paper, we describe our efforts to achieve human parity on large-scale datasets for a Chinese-English news translation task. We  address a number of limitations of the current NMT paradigm. Our contributions are:
\begin{itemize}
\item We utilize the duality of the translation problem to allow the model to learn from both source-to-target and target-to-source translations. Simultaneously this allows us to learn from both supervised and unsupervised source and target data. This will be described in Section \ref{sec.duality}. Specifically, we utilize a generic Dual Learning approach ~\cite{dualNMT,DSL,dualInfer}   (Section \ref{sec.dl}), and introduce a joint training algorithm to enhance the effect of monolingual source and target data by iteratively boosting the  source-to-target and target-to-source translation models in a unified framework (Section \ref{sec-s2t-t2s}).

\item NMT systems decode auto-regressively from left-to-right, which means that during sequential generation of the output, previous errors will be amplified and may mislead subsequent generation. This is only partially remedied by beam search. We propose two approaches to alleviate this problem: Deliberation Networks~\cite{delibnet} is a method to refine the translation based on two-pass decoding (Section \ref{sec.dn}); and a new training objective over two Kullback-Leibler (KL) divergence regularization terms encourages agreement between left-to-right and right-to-left decoding results (Section \ref{sec-l2r-r2l}). 

\item Since NMT is very vulnerable to noisy training data, rare occurrences in the data, and the training data quality in general \cite{noisyNMT}. We discuss our approaches for data selection and filtering, including a cross-lingual sentence representation, in Section \ref{sec.ds}. 

\item Finally, we find that our systems are quite complementary, and can therefore benefit greatly from system combination, ultimately attaining human parity. See section \ref{sec.sc}.
\end{itemize}

In this work, we interchangeably use source-to-target and (Zh$\rightarrow$En) to denote Chinese-to-English; target-to-source and (En$\rightarrow$Zh) to denote English-to-Chinese.
\subsection{Exploiting the Dual Nature of Translation}
\label{sec.duality}
We leverage  the duality of the translation problem to allow the model to learn from both source-to-target and target-to-source translations. We explore the translation duality using two approaches: Dual Learning \ref{sec.dl} and Joint Training \ref{sec-s2t-t2s}
\subsubsection{Dual Learning for NMT}
\label{sec.dl}
Dual learning~\cite{dualNMT,DSL,dualInfer}, a recently proposed learning paradigm, tries to achieve the co-growth of machine learning models in two dual tasks, such as image classification vs. image generation, speech recognition vs. text-to-speech, and Chinese to English vs. English to Chinese translation. In dual learning, the two parallel models (referred to as the \emph{primal model} and the \emph{dual model}) enhance each other by leveraging primal-dual structure in order to learn from unlabeled data or regularize the learning from labeled data. Ever since dual learning was proposed, it has been successfully applied to various real-world problems such as question answering~\cite{tang2017question}, image classification~\cite{DSL}, image segmentation~\cite{deepdual}, image to image translation~\cite{dualgan,cyclegan,cdgan}, face attribute manipulation~\cite{face}, and machine translation~\cite{dualNMT,wang2018dt,unsupervisedNMT,artetxe2018unsupervised}.

In this work, to achieve strong machine translation performance, we combine two different dual learning methods that respectively enhance the usage of monolingual and bilingual training data. We set the Chinese to English (Zh$\rightarrow$En) translation model as the primal model and the English to Chinese (En$\rightarrow$Zh) model as the dual model, respectively denoted as $p(\mathbf{y}|\mathbf{x};\theta_{x\rightarrow y})$ and $p(\mathbf{x}|\mathbf{y};\theta_{y\rightarrow x})$.

\begin{itemize}
\item \emph{Dual unsupervised learning} (DUL)~\cite{dualNMT}. To enhance the Zh$\rightarrow$En translation quality, DUL efficiently leverages a monolingual Chinese corpus based on additional supervision signals from the dual En$\rightarrow$Zh model. Concretely speaking, for a monolingual Chinese sentence $\mathbf{x}$, an English translation $\mathbf{y}$ is sampled using the primal model $p(\cdot|\mathbf{x};\theta_{x\rightarrow y})$; starting from $\mathbf{y}$, we use the dual model $p(\cdot|\mathbf{y};\theta_{y\rightarrow x})$ to compute the log-likelihood $\log p(\mathbf{x}|\mathbf{y};\theta_{y\rightarrow x})$ of reconstructing $\mathbf{x}$ from $\mathbf{y}$ and treat it as the reward of taking action $\mathbf{y}$ at state $\mathbf{x}$. We would like to maximize the expected reconstruction log-likelihood when iterating over all possible translation $\mathbf{y}$ for $\mathbf{x}$, shown as:

\begin{equation}
\mathcal{L}(\mathbf{x};\theta_{x\rightarrow y})=E_{\mathbf{y}\sim p(\cdot|\mathbf{x};\theta_{x\rightarrow y})}\bigl\{\log p(\mathbf{x}|\mathbf{y};\theta_{y\rightarrow x})\bigr\}=\sum_{\mathbf{y}}p(\mathbf{y}|\mathbf{x};\theta_{x\rightarrow y})\log p(\mathbf{x}|\mathbf{y};\theta_{y\rightarrow x})
\end{equation}

Taking the gradient of $\mathcal{L}(\mathbf{x};\theta_{x\rightarrow y})$ with respect to $\theta_{x\rightarrow y}$, we obtain:
\begin{equation}
\begin{aligned}
\frac{\partial \mathcal{L}(\mathbf{x};\theta_{x\rightarrow y})}{\partial \theta_{x\rightarrow y}}&=\sum_{\mathbf{y}}\frac{\partial p(\mathbf{y}|\mathbf{x};\theta_{x\rightarrow y})}{\partial \theta_{x\rightarrow y}}\log p(\mathbf{x}|\mathbf{y};\theta_{y\rightarrow x})\\
&= \sum_{\mathbf{y}}p(\mathbf{y}|\mathbf{x};\theta_{x\rightarrow y})\frac{\partial \log p(\mathbf{y}|\mathbf{x};\theta_{x\rightarrow y})}{\partial \theta_{x\rightarrow y}}\log p(\mathbf{x}|\mathbf{y};\theta_{y\rightarrow x})
\end{aligned}
\end{equation}

Since summing over all possible $\mathbf{y}$ in the above equation is computationally intractable, we use Monte Carlo sampling to approximate the above expectation:
\begin{equation}
\frac{\partial \mathcal{L}(\mathbf{x};\theta_{x\rightarrow y})}{\partial \theta_{x\rightarrow y}}\approx \frac{\partial \log p(\mathbf{y}'|\mathbf{x};\theta_{x\rightarrow y})}{\partial \theta_{x\rightarrow y}}\log p(\mathbf{x}|\mathbf{y}';\theta_{y\rightarrow x}),
\end{equation} where $\mathbf{y}'$ is a sampled translation from the primal model $p(\cdot|\mathbf{x};\theta_{x\rightarrow y})$. 


The approximated gradient is used to update the primal model parameters $\theta_{x\rightarrow y}$. 
Note that the parameters of the dual model $\theta_{y\rightarrow x}$ can be updated using a monolingual English corpus in a similar way by maximizing the reconstruction likelihood from possible Chinese translations. 

\item \emph{Dual supervised learning} (DSL)~\cite{DSL}. Unlike DUL, which aims to effectively leverage monolingual data, DSL is an approach to better utilize bilingual training data by enhancing probabilistic correlations within the two models. The idea of DSL is to force the joint probability consistency within primal model and dual model. Specifically, for a bilingual sentence pair $(\mathbf{x},\mathbf{y})$, ideally we have $p(\mathbf{x},\mathbf{y})=p(\mathbf{x})p(\mathbf{y}|\mathbf{x})=p(\mathbf{y})p(\mathbf{x}|\mathbf{y})$. However, if the two models are trained separately, it is hard for them to satisfy $p(\mathbf{x})p(\mathbf{y}|\mathbf{x})=p(\mathbf{y})p(\mathbf{x}|\mathbf{y})$. Therefore, when applied in neural machine translation, DSL conducts joint training of the two models and introduces an additional loss term on the parallel data $(\mathbf{x},\mathbf{y})$ for regularization:
\begin{equation}
\mathcal{L}_{DSL}=(\log \hat{p}(\mathbf{x})+\log p(\mathbf{y}|\mathbf{x};\theta_{x\rightarrow y})-\log \hat{p}(\mathbf{y})-\log p(\mathbf{x}|\mathbf{y};\theta_{y\rightarrow x}))^2,
\end{equation}
where $\hat{p}(\mathbf{x})$ and $\hat{p}(\mathbf{y})$ are empirical marginal distributions induced by the training data. In our experiments, they are the output scores of two language models respectively trained on Chinese and English corpus containing both bilingual and monolingual data.
\end{itemize}

In our architecture, both DUL and DSL are used in model training, both of which are applied to the monolingual and bilingual training corpora.

\subsubsection{Joint Training of Source-to-Target and Target-to-Source Models}
\label{sec-s2t-t2s}

Back translation \cite{sennrich2015improving} augments relatively scarce parallel data with plentiful monolingual data, allowing us to train source-to-target (S2T) models with the help of target-to-source (T2S) models.  
Specifically, given a set of sentences $\{\mathbf{y}^{(t)}\}$ in the target language, a pre-constructed T2S translation system is used to automatically generate translations $\{\mathbf{x}^{(t)}\}$ in the source language. These synthetic sentence pairs $\{(\mathbf{x}^{(t)}, \mathbf{y}^{(t)})\}$ are combined with the original bilingual data when training the S2T NMT model.
In order to leverage both source and target language monolingual data, and also let S2T and T2S models help each other,
we leverage the joint training method described in \cite{Joint_S2T_T2S} to optimize them by extending the back-translation method. The joint training method uses the  monolingual data and updates NMT models through several iterations.

Given parallel corpus $D=\{(\mathbf{x}^{(n)},\mathbf{y}^{(n)})\}_{n=1}^{N}$ and target monolingual corpus  $Y = \{\mathbf{y}^{(t)}\}_{t=1}^{T}$, a semi-supervised training objective is used to jointly maximize the likelihood of both bilingual data and monolingual data:
\begin{equation}
\begin{aligned}
\mathcal{L}^*(\theta_{x\rightarrow y}) &= \sum_{n=1}^N \log p(\mathbf{y}^{(n)}|\mathbf{x}^{(n)}) + \sum_{t=1}^T\log p(\mathbf{y}^{(t)})
\end{aligned}
\label{equ:semi-loss-s2t}
\end{equation}

By introducing $\mathbf{x}$ as the latent variable representing the source translation of target sentence $\mathbf{y}^{(t)}$, Equation \ref{equ:semi-loss-s2t} can be optimized in an EM framework, with the help of a T2S translation model:
\begin{equation}
\begin{aligned}
\mathcal{L}(\theta_{x\rightarrow y}) = & \sum_{n=1}^N \log p(\mathbf{y}^{(n)}|\mathbf{x}^{(n)})  + \sum_{t=1}^T\sum_{\mathbf{x}} p(\mathbf{x}|\mathbf{y}^{(t)}) \log p(\mathbf{y}^{(t)}|\mathbf{x}) 
\end{aligned}
\label{equ:semi-loss-opt1}
\end{equation}

Similarly, we can optimize the T2S translation model with the help of S2T translation model as follows:
\begin{equation}
\begin{aligned}
\mathcal{L}(\theta_{y\rightarrow x}) = & \sum_{n=1}^N \log p(\mathbf{x}^{(n)}|\mathbf{y}^{(n)})  + \sum_{s=1}^S\sum_{\mathbf{y}} p(\mathbf{y}|\mathbf{x}^{(s)}) \log p(\mathbf{x}^{(s)}|\mathbf{y})
\end{aligned}
\label{equ:semi-loss-opt2}
\end{equation}

\begin{figure}
    \centering
    \begin{minipage}[t]{0.48\textwidth}\centering%
        \includegraphics[width=\linewidth]{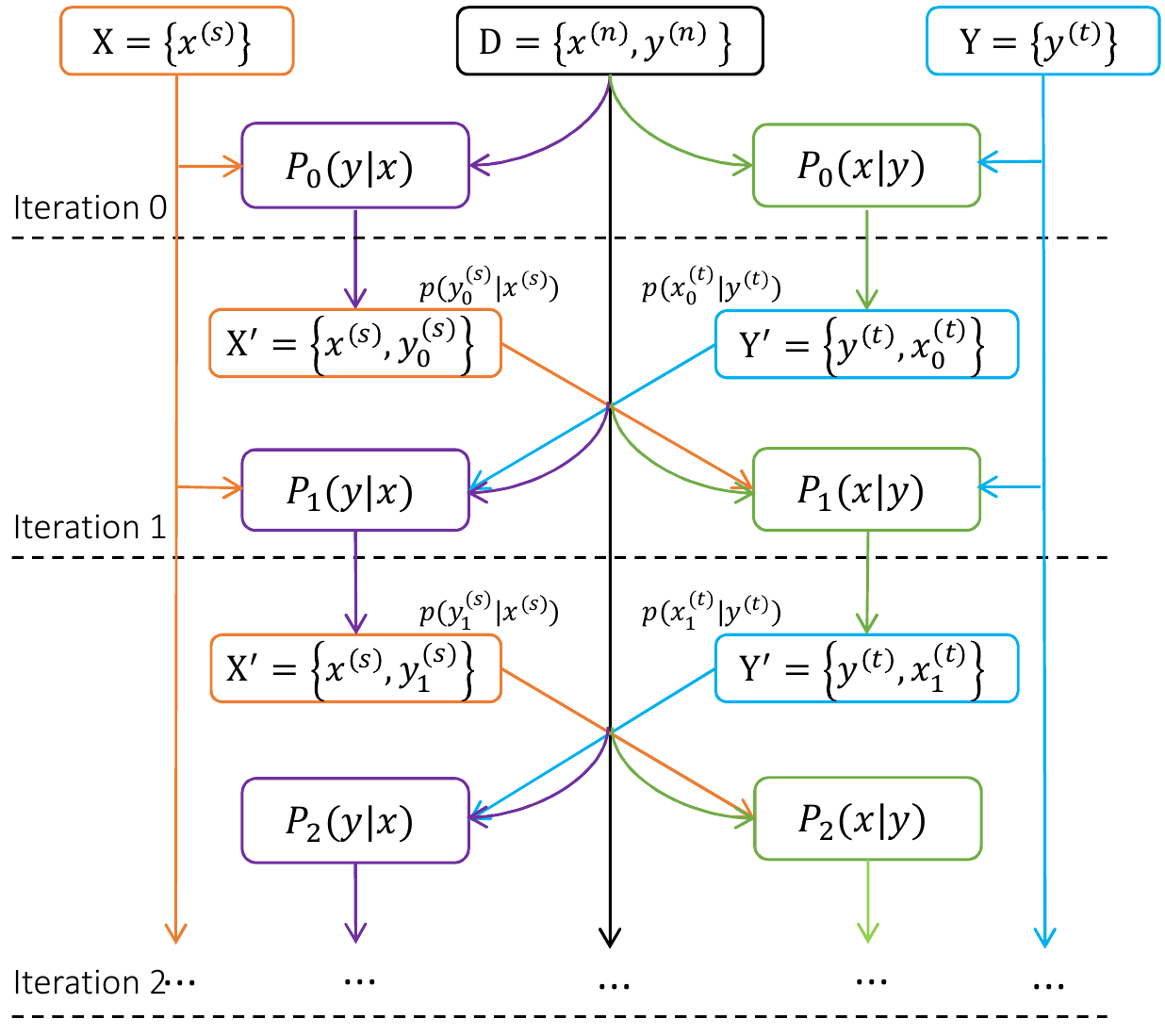}
  		\caption{Illustration of joint training: S2T $p(\mathbf{y}| \mathbf{x})$ and T2S $p(\mathbf{x}|\mathbf{y})$}
  \label{fig:S2TT2S}
    \end{minipage}\hfill
    \begin{minipage}[t]{0.48\textwidth}\centering%
  		\includegraphics[width=\textwidth]{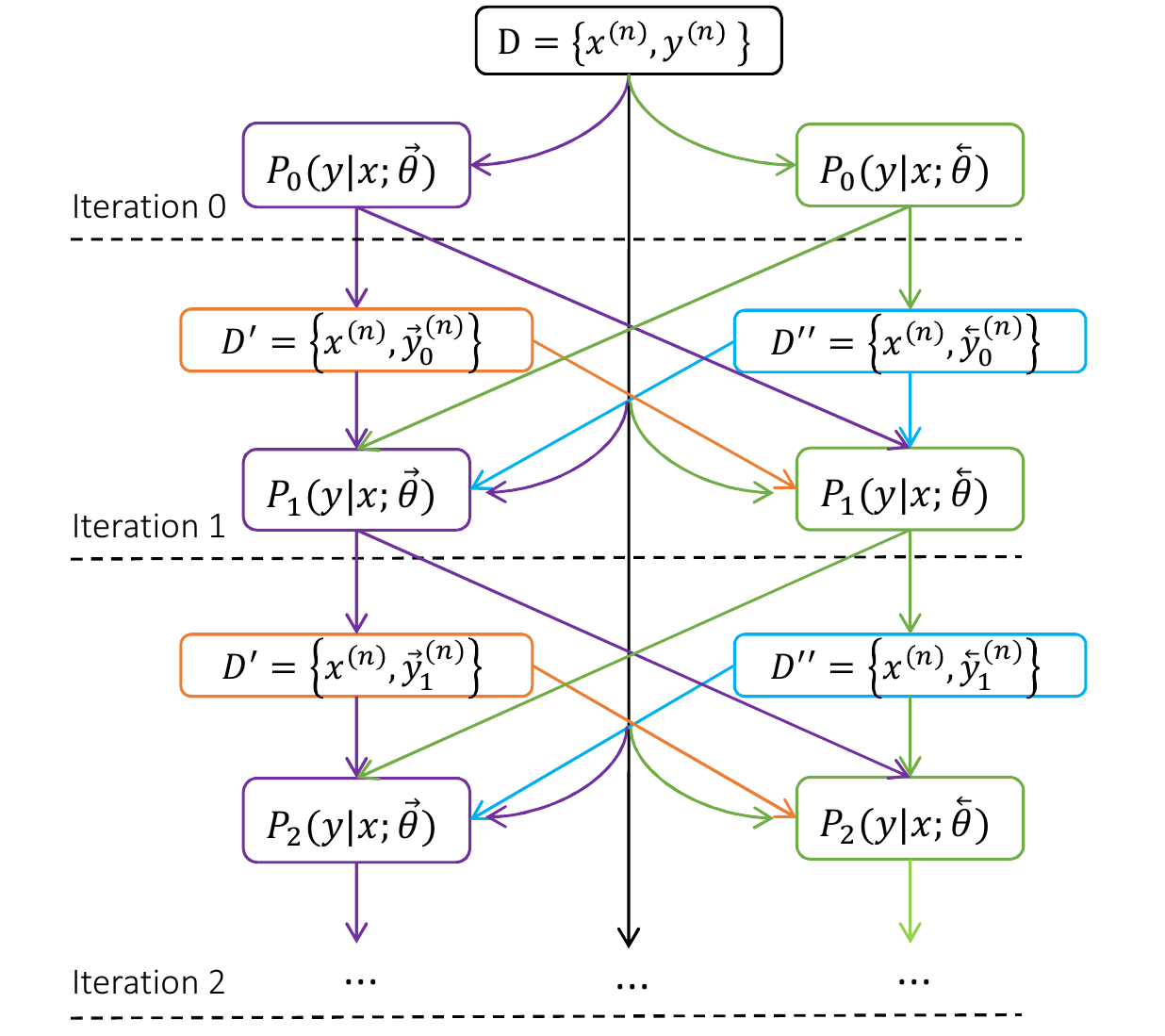}
		\caption{Illustration of agreement regularization: L2R $p(\mathbf{y}|\mathbf{x};\protect\overrightarrow{\theta})$ and R2L $p(\mathbf{y}|\mathbf{x};\protect\overleftarrow{\theta})$}
		\label{fig:L2RR2L}
	\end{minipage}
\end{figure}

As we can find from Equation \ref{equ:semi-loss-opt1} and \ref{equ:semi-loss-opt2},  model $p(\mathbf{y}|\mathbf{x})$ and $p(\mathbf{x}|\mathbf{y})$ serve as each other's pseudo-training data generator: $p(\mathbf{x}|\mathbf{y})$ is used to translate $Y$ into $X$ for $p(\mathbf{y}|\mathbf{x})$, while $p(\mathbf{y}|\mathbf{x})$ is used to translate $X$ to $Y$ for $p(\mathbf{x}|\mathbf{y})$.  
The joint training process is illustrated in Figure~\ref{fig:S2TT2S}. Before the first iteration starts, two initial translation models $p_0(\mathbf{y}|\mathbf{x})$ and $p_0(\mathbf{x}|\mathbf{y})$ are pre-trained with parallel data $D = \{(\mathbf{x}^{(n)},\mathbf{y}^{(n)})\}$. This step is denoted as iteration 0 for sake of consistency. In iteration 1, two NMT systems $p_0(\mathbf{y}|\mathbf{x})$ and $p_0(\mathbf{x}|\mathbf{y})$ are used to translate monolingual data $X=\{\mathbf{x}^{(s)}\}$ and $Y=\{\mathbf{y}^{(t)}\}$, which creates two synthetic training data sets $X'=\{\mathbf{x}^{(s)}, \mathbf{y}_0^{(s)}\}$ and $Y'=\{\mathbf{y}^{(t)}, \mathbf{x}_0^{(t)}\}$. 
Models $p_1(\mathbf{y}|\mathbf{x})$ and $p_1(\mathbf{x}|\mathbf{y})$ are then trained on this augmented training data by combining $Y'$ and $X'$ with parallel data $D$.
It is worth noting that $n$-best translations are used, and the selected translations are weighted with the translation probabilities given by the NMT model, so that the negative impact of noisy translations can be minimized. In iteration 2, the above process is repeated, and the synthetic training data are re-generated with the updated NMT models $p_1(\mathbf{y}|\mathbf{x})$ and $p_1(\mathbf{x}|\mathbf{y})$, which are presumably more accurate. The learned NMT models $p_2(\mathbf{y}|\mathbf{x})$ and $p_2(\mathbf{x}|\mathbf{y})$ are also expected to improve with better pseudo-training data. The training process continues until the performance on a development data set is no longer improved. 

\subsection{Beyond the Left-to-Right Bias}
\label{sec.delb}

Current NMT systems suffer from the {\em exposure bias} problem \cite{bengio2015scheduled}. Exposure bias refers to the problem that during sequential generation of output, previous errors will be amplified and mislead subsequent generation. We address this limitation in two ways: a two-pass decoding (Deliberation Networks) \ref{sec.dn} and Agreement Regularization \ref{sec-l2r-r2l}.

\subsubsection{Deliberation Networks}
\label{sec.dn}
\begin{figure}[!htpb]
                \centering
                \includegraphics[width=0.9\linewidth]{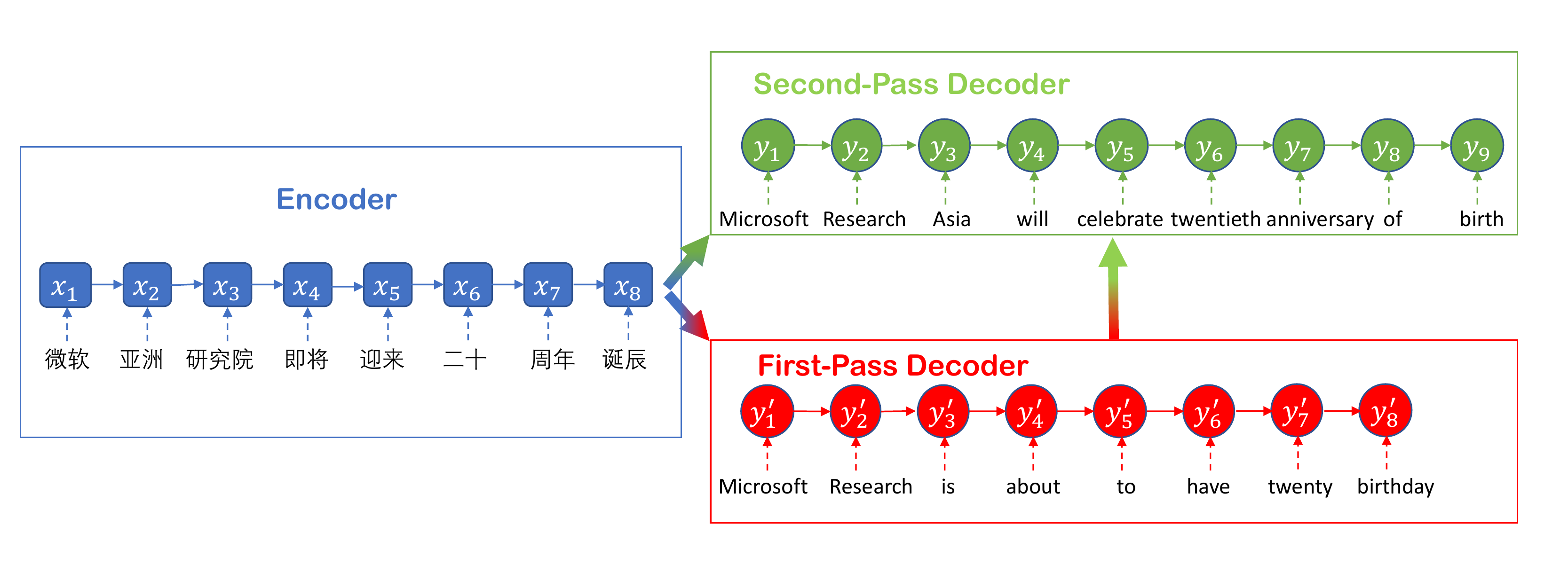}
                \caption{An example showing the decoding process of deliberation network.}
                \label{fig:delib}
\end{figure}

Classical neural machine translation models generate a translation word by word from left to right, all in one pass. This is very different from human behavior such as, for instance, while writing articles or papers. When writing papers, usually we create a first draft, then we revisit the draft in its full context, further polishing each word (or phrase/sentence/paragraph) based on both its left-side context and right-side context. In contrast, in neural machine translation, decoding in only one pass makes the output of the $t$-th word $y_t$ dependent on the source-side sentence $\mathbf{x}$ and its left context only (i.e., already generated tokens $\{y_1,\cdots,y_{t-1}\}$), without any opportunity to look into the future. Inspired by the human writing process, Deliberation Networks~\cite{delibnet} try to overcome this drawback by decoding using a two-pass process with two decoders as illustrated in Fig.~\ref{fig:delib}. The first-pass decoder outputs an initial translation as a draft. The second-pass decoder polishes this draft into a final translation. The draft translation output from the first pass decoder contains global information that enlarges the receptive field of decoding each token $y_t$ in the second-pass decoding process, and thus breaks the limitation of only looking to the left-hand side. 

\begin{figure}[!htpb]
	\centering
	\includegraphics[width=0.9\linewidth]{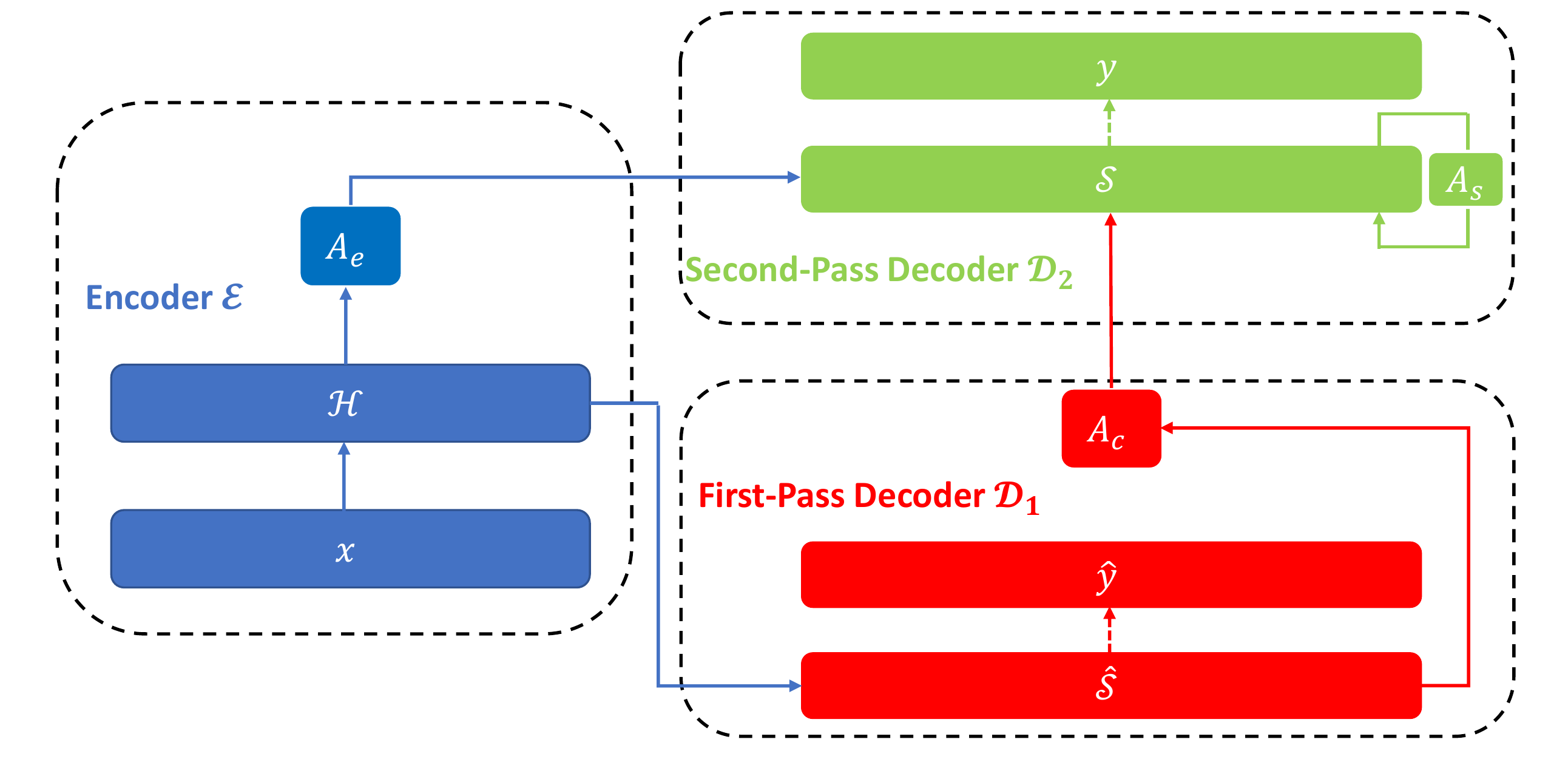}
	\caption{Deliberation network: Blue, red and green parts indicate encoder $\mathcal{E}$, first-pass decoder $\mathcal{D}_1$ and second-pass decoder $\mathcal{D}_2$ respectively. Solid lines represent the information flow via attention model. The self attention model within $\mathcal{E}$ and the $\mathcal{E}$-to-$\mathcal{D}_1$ attention model are omitted for readability.}
	\label{fig:delib_detail}
\end{figure}

The detailed model architecture, with a deliberation network built on top of Transformer, is shown in Fig.~\ref{fig:delib_detail}. As in standard Transformer, both the encoder $\mathcal{E}$ and the first-pass decoder $\mathcal{D}_1$ contain several stacked layers connected via a self attention mechanism.
Specifically, the encoder assigns to each of the $T_s$ source words a representation based on its original embedding and contextual information gathered from other positions. We denote this sequence of top-layer state vectors $h_{1:T_s}$ as $\mathcal{H}$.
%
%
The encoder $\mathcal{E}$ reads the source sentence $x$ and outputs a sequence of hidden states $\mathcal{H}\,=\,h_{1: T_s}$ via self attention. The first-pass decoder $\mathcal{D}_1$ takes $\mathcal{H}$ as inputs, conducts the first round decoding and obtains the first-pass translation sentence $\hat{\mathbf{y}}$ as well as the hidden states before softmax denoted as $\hat{\mathcal{S}}$. The second-pass decoder $\mathcal{D}_2$ also contains several stacked layers, but is significantly different from $\mathcal{D}_1$ in that $\mathcal{D}_2$ takes the hidden states output by both $\mathcal{E}$ and $\mathcal{D}_1$ as inputs. Specifically, denoting the output of the $i$th layer in $\mathcal{D}_2$ as $s^i$, we have $s^i=A_e(\mathcal{H}, s^{i-1})+A_c(\hat{\mathcal{S}},s^{i-1}) + A_s(s^{i-1})$, where $A_e$ and $A_c$ are the multi-head attention mechanism~\cite{vaswani2017attention} connecting $\mathcal{D}_2$ respectively with $\mathcal{E}$ and $\mathcal{D}_1$, and $A_s$ is the self attention mechanism within $\mathcal{D}_2$ operating on $s^{i-1}$. It is easily observed that the last translation result $y$ is dependent on the first translation sentence $\hat{\mathbf{y}}$, since we feed the outputs of the first-pass decoder $\mathcal{D}_1$ into the second-pass decoder $\mathcal{D}_2$. In this way we obtain global information on the target side, thereby allowing us to look at right context in sentence generation. Policy gradient algorithms are used to jointly optimize the parameters of the three parts.

The combination of dual learning and deliberation networks takes place as follows: First, we train the  Zh$\rightarrow$En and En$\rightarrow$Zh Transformer models using both DUL and DSL. Then, for a target side monolingual sentence $\mathbf{y}$, the existing En$\rightarrow$Zh model is used to translate it into Chinese sentence $\mathbf{x'}$. Afterwards, we treat $(X',Y)$ as pseudo bilingual data and add it into the bilingual data corpus. The enlarged bilingual corpus is then used to train the deliberation network as described above. In deliberation network training, we use the Zh$\rightarrow$En model obtained in the first step to initialize the encoder and first-pass decoder. 

\subsubsection{Agreement Regularization of Left-to-Right and Right-to-Left Models}
\label{sec-l2r-r2l}
An alternative way of addressing exposure bias is to leverage the fact that
%
unsatisfactory translations with bad suffixes generated by a left-to-right (L2R) model usually have low prediction scores under a right-to-left (R2L) model.
In the R2L model, if bad suffixes are fed as inputs to the decoder first, this will lead to corrupted hidden states, therefore good prefixes reached later will be given considerably lower prediction probabilities. This signal given by the R2L model can be leveraged to alleviate the exposure bias problem of the L2R model and vice versa.

To train the L2R model, two Kullback-Leibler (KL) divergence regularization terms are introduced into the maximum-likelihood training objective, as shown in
\begin{equation}
\begin{aligned}
\mathcal{L}(\overrightarrow{\theta}) = &\sum_{n=1}^{N}  \log{ p(\mathbf{y}^{(n)}| \mathbf{x}^{(n)};\overrightarrow{\theta} )} 
 - \lambda \sum_{n=1}^{N} \text{KL}( p(\mathbf{y}|\mathbf{x}^{(n)}; \overleftarrow{\theta} ) || p(\mathbf{y}|\mathbf{x}^{(n)}; \overrightarrow{\theta} ) ) \\
& - \lambda \sum_{n=1}^{N} \text{KL}( p(\mathbf{y}|\mathbf{x}^{(n)}; \overrightarrow{\theta} ) || p(\mathbf{y}|\mathbf{x}^{(n)}; \overleftarrow{\theta} ) ) 
\end{aligned}
\label{equ:regularization}
\end{equation}
With a simple mathematic calculation and proper approximation, we can get the parameter gradients for L2R model as follows:
\begin{equation}
\begin{aligned}
\frac{ \partial \mathcal{L}(\overrightarrow{\theta}) }{ \partial \overrightarrow{\theta}} & = \sum_{n=1}^{N} \frac{ \partial  \log{ p(\mathbf{y}^{(n)}| \mathbf{x}^{(n)};\overrightarrow{\theta} )} }{ \partial \overrightarrow{\theta} }  
 + \lambda \sum_{n=1}^{N} \sum_{\mathbf{y} \sim p(\cdot|\mathbf{x}^{(n)}; \overleftarrow{\theta} )} \frac{\partial \log p(\mathbf{y}|\mathbf{x}^{(n)}; \overrightarrow{\theta} ) }{\partial \overrightarrow{\theta}} \\
& + \lambda \sum_{n=1}^{N} \sum_{\mathbf{y} \sim p(\cdot|\mathbf{x}^{(n)}; \overrightarrow{\theta} )}
 \left( \log \frac{p(\mathbf{y}|\mathbf{x}^{(n)}; \overleftarrow{\theta} )}{ p(\mathbf{y}|\mathbf{x}^{(n)}; \overrightarrow{\theta} )} 
\frac{\partial \log p(\mathbf{y}|\mathbf{x}^{(n)}; \overrightarrow{\theta} ) }{\partial \overrightarrow{\theta}} \right)
\end{aligned}
\label{equ:gradient_l2r}
\end{equation}
The first part tries to maximize the log likelihood of the bilingual training corpus. The second part maximizes the log likelihood of the "pseudo corpus"  constructed by the R2L model. The third part maximizes a weighted log likelihood of another pseudo corpus generated by the L2R model itself with a weight of ($\log (p(\mathbf{y}|\mathbf{x}^{(n)}; \overleftarrow{\theta} ) / p(\mathbf{y}|\mathbf{x}^{(n)}; \overrightarrow{\theta}))$) which penalizes the samples where the L2R and R2L models do not agree. We find that the R2L model plays the role of an auxiliary system which provides a pseudo corpus in the second part and calculates the weight in the third part.

Similarly, we can get corresponding parameter gradients for the R2L model by introducing two KL divergence regularization terms, as follows:

\begin{equation}
\begin{aligned}
\frac{ \partial  \mathcal{L}(\overleftarrow{\theta}) }{ \partial \overleftarrow{\theta}} & = \sum_{n=1}^{N} \frac{ \partial  \log{ p(\mathbf{y}^{(n)}| \mathbf{x}^{(n)};\overleftarrow{\theta} )} }{ \partial \overleftarrow{\theta} }  
 + \lambda \sum_{n=1}^{N} \sum_{\mathbf{y} \sim p(\cdot|\mathbf{x}^{(n)}; \overrightarrow{\theta} )} \frac{\partial \log p(\mathbf{y}|\mathbf{x}^{(n)}; \overleftarrow{\theta} ) }{\partial \overleftarrow{\theta}} \\
& + \lambda \sum_{n=1}^{N} \sum_{\mathbf{y} \sim p(\cdot|\mathbf{x}^{(n)}; \overleftarrow{\theta} )}
 \left( \log \frac{p(\mathbf{y}|\mathbf{x}^{(n)}; \overrightarrow{\theta} )}{ p(\mathbf{y}|\mathbf{x}^{(n)}; \overleftarrow{\theta} )} 
\frac{\partial \log p(\mathbf{y}|\mathbf{x}^{(n)}; \overleftarrow{\theta} ) }{\partial \overleftarrow{\theta}} \right)
\end{aligned}
\label{equ:gradient_r2l}
\end{equation}

With the help of the R2L model, the L2R model can be enhanced using Equation \ref{equ:gradient_l2r}. With the enhanced L2R model, a better pseudo corpus and more accurate weights can be leveraged to improve the performance of the R2L model with Equation \ref{equ:gradient_r2l}, while simultaneously this better R2L model can be reused to improve the L2R model. In such a way, L2R and R2L models can mutually boost each other as illustrated in Figure \ref{fig:L2RR2L}. The training process continues until the performance on a development data set is no further improving.

\begin{algorithm}[t]
\caption{Unified Joint Training Algorithm}
\label{alg:unified_joint_training}
\hspace*{\algorithmicindent} \textbf{Input:} Bilingual Data $ D =\{ (\mathbf{x}^{(n)}, \mathbf{y}^{(n)}) \}_{n=1}^{N}$,  Source and Target Monolingual Corpora $X=\{\mathbf{x}^{(s)}\}_{s=1}^{S}$ and $Y=\{\mathbf{y}^{(t)}\}_{t=1}^{T}$;  \\
\hspace*{\algorithmicindent} \textbf{Output:} S2T-L2R Model $ p(\overrightarrow{\mathbf{y}}|\mathbf{x}) $, S2T-R2L Model $ p(\overleftarrow{\mathbf{y}}|\mathbf{x}) $, T2S-L2R Model $ p(\overrightarrow{\mathbf{x}}|\mathbf{y}) $ and T2S-R2L Model $ p(\overleftarrow{\mathbf{x}}|\mathbf{y}) $;
\begin{algorithmic}[1]
\Procedure{training process}{}
\State Pre-train four models with maximum likelihood on parallel corpora $ D =\{ (\mathbf{x}^{(n)}, \mathbf{y}^{(n)}) \}_{n=1}^{N}$;
\While{Not Converged}
\State Build weighted pseudo-parallel corpora $Y'=\{ (\mathbf{x}^{(t)}, \mathbf{y}^{(t)}) \}_{t=1}^{T}$ with \( p(\overrightarrow{\mathbf{x}}|\mathbf{y}) \) using monolingual data $Y=\{\mathbf{y}^{(t)}\}_{t=1}^{T}$ as shown in Figure \ref{fig:S2TT2S}.
\State Update $ P(\overleftarrow{\mathbf{y}}|\mathbf{x}) $ and $ p(\overrightarrow{\mathbf{y}}|\mathbf{x}) $  as shown in Figure \ref{fig:L2RR2L}, with original data $ D =\{ (\mathbf{x}^{(n)}, \mathbf{y}^{(n)}) \}_{n=1}^{N}$ and synthetic data $Y'=\{ (\mathbf{x}^{(t)}, \mathbf{y}^{(t)}) \}_{t=1}^{T}$.
\State Build weighted pseudo-parallel corpora $\mathbf{x}'=\{ (\mathbf{x}^{(s)}, \mathbf{y}^{(s)}) \}_{s=1}^{S}$ with \( p(\overrightarrow{\mathbf{y}}|\mathbf{x}) \) using monolingual data $X=\{\mathbf{x}^{(s)}\}_{s=1}^{S}$ as introduced in Figure \ref{fig:S2TT2S}. 
\State Update $ p(\overleftarrow{\mathbf{x}}|\mathbf{y}) $ and $ P(\overrightarrow{\mathbf{x}}|\mathbf{y}) $ as shown in Figure \ref{fig:L2RR2L}, with original data $ D =\{ (\mathbf{x}^{(n)}, \mathbf{y}^{(n)}) \}_{n=1}^{N}$ and synthetic data $X'=\{ (\mathbf{\mathbf{x}}^{(s)}, \mathbf{y}^{(s)}) \}_{s=1}^{S}$.
\EndWhile
\EndProcedure
\end{algorithmic}
\end{algorithm}

Since both the source and target sentences can be generated from left to right and from right to left, we can have a total of four systems, two source to target models: S2T-L2R (target sentence is generated from left to right), S2T-R2L (target sentence is generated from right to left), and two target to source models: T2S-L2R (source sentence is generated from left to right), T2S-R2L (source sentence is generated from right to left). Using the agreement regularization method described above, these four models can be optimized in a unified joint training framework, as shown in Algorithm \ref{alg:unified_joint_training}. With the joint training method, a weighted pseudo corpus is generated by T2S-L2R model and used to train two S2T models (S2T-L2R and S2T-R2L) with the help of agreement regularization. The enhanced S2T-L2R model is then used to build another weighted pseudo corpus to train two T2S models. These four systems boost each other until convergence is reached.


\subsection{Data Selection and Filtering}
\label{sec.ds}
Though NMT systems require huge amounts of training data,  not all data are equally useful for training the systems.
NMT systems are more vulnerable to noisy training data, rare occurrences in the data, and the training data quality in general.  We are trying to tackle two different problems: selecting data relevant  to the task and removing noisy data. Out-of-domain and noisy data are distinct problems and may harm the system in different ways. Many studies have highlighted the bad impact of noisy data on MT, such as \cite{noisyNMT}. Even small amounts of noisy data can have very bad effects since NMT models tend to assign high probabilities to rare events. Noise in data can take several forms, including totally incorrect translations, partial translations, inaccurate or machine translated data, wrong source or target language, or source copied to the target. We use  features from word alignment to filter out the very noisy data, similar to the approach in \cite{datagen}. However, data that is less egregiously noisy  represents a bigger problem since it is harder to recognize.

The de-facto standard method for data selection for SMT is \cite{Moorelewis} and \cite{Axelrod}. Unfortunately it has not proved as useful for NMT; while it reduces the training data it does not lead to improvements in system quality \cite{ds_nmt}.  We propose a new approach that tackles both problems at once: filtering noisy data and selecting relevant data. Our approach centers on first learning a bilingual sentence vector representation where sentences in both languages are mapped into the same space. After learning this representation, we use it for both  filtering noisy data and selecting relevant data.

To learn our sentence representation we train a unified bilingual NMT system similar to \cite{zoph2016transfer} that can translate between Chinese and English in both directions. We train this on a selected subset of the data  that is known to be of good quality and in the relevant domain. Building the model with such relevant data has two advantages. First: it helps the representation to be similar to the cleaner data; second: relevant sentences would have better representation than irrelevant ones. Therefore we would achieve both data cleaning and relevant data selection objectives.

Recent progress in  multi-lingual NMT i.e.\ \cite{johnson2016google} and \cite{unimt} shows that these models are able to represent multiple languages in the same space. However, we don't use language markers because we want to force the model to learn similar representations for both Chinese and English. Given this bilingual system, for any sentence in Chinese or English we can run the encoder part of the system to get a contextual vector representation for each word of a sentence. This is the vector from the last encoder layer, normally used as input to the attention model. We represent each sentence vector as the mean of the word-level contextual vectors.

Specifically, the encoder assigns to each of the $T_s$ source words a representation based on its original embedding and contextual information gathered from other positions. We denote this set of  top-layer state vectors as $h_{1:T_s}$:
%
\begin{equation}
h_{1: T_s} = f^{\text{enc}}\bigl(E(x_1), ..., E(x_{T_s})) 
\label{eq.encoder}
\end{equation}
where $E^I \in \mathbb{R}^{V \times d}$ is a look-up table of joint source and target embeddings, assigning each individual word a unique embedding vector.

If $h^{\mathrm{enc}}_{1:T_s}$ denotes the encoder's top layer's output sequence,
the sentence-vector representation $S_{sv}$ of a given sentence $S$ of length $T_s$ is:
%
%
%
\begin{equation}
	\begin{split}
	S_{sv}=\sum_{\ell=1}^{T_s}h^{\mathrm{enc}}_{\ell}
	\end{split}
	 \label{eq.sentvec} 
\end{equation}

A similarity measure  $\mathrm{SIM}_{ST}$ between any two given sentences $S$ and $T$, regardless of their languages, can be represented as the cosine similarity between their corresponding sentences vectors:

\begin{equation}
	\mathrm{SIM}_{ST} = \dfrac{S_{sv}\cdot T_{sv}} {\vert{S_{sv}}\vert \vert{T_{sv}}\vert}	 \label{eq.sentsim} 
\end{equation}

We train an RNN encoder-decoder system similar to \cite{wu2016google} with 4 encoder layers  with the first layer being bidirectional and 4 decoder layers and an attention model.   After training the model, we run the encoder part only.  Each resulting word context vector is composed of an 1024 dimension  vector; therefore the sentence vector ($S_{sv}$) representation is of the same size.

For each sentence in the parallel training corpus, we measure the cross-lingual  similarity between source and target sentences as in Equation \ref{eq.sentsim}. We reject sentences with  similarity below a specified threshold. This approach enables us to drastically reduce the training data while significantly improving the accuracy. Since we use a model trained on relevant data, this data selection technique can serve a dual purpose by filtering noisy data as well as selecting relevant data.


\subsection{System Combination and Re-ranking}
\label{sec.sc}
In order to combine the systems described above, we combine n-best hypotheses from all systems and then train a re-ranker using k-best MIRA on the validation set. K-best MIRA \cite{kb-mira}  is a version of MIRA (a margin-based classification algorithm) that  works with a batch tuning  to learn a re-ranker for the k-best hypothesis.

The features we use for re-ranking are:
\begin{itemize}
\item $SYS_{Score}$: Original System Score and identifier. 
\item $LM_{Score}$: 5-gram language model trained on English news crawled data of 2015 and 2016. 
\item $R2L_{score}$: R2L system re-scoring. A system trained on Chinese source and reversed English target; the system is used to score each hypothesis.
\item $E2Z_{score}$ : English-to-Chinese  system re-scoring. A system trained on English to Chinese is used to score each hypothesis. .
\item $ST_{SV}$ : Cross-lingual sentence similarity between source and the hypothesis as described in Section \ref{sec.ds}. 
\item $R2L_{SV}$: R2L sentence vector similarity: the best hypothesis from the R2L system is compared to each n-best hypothesis and used to generate a sentence similarity score based on sentence vector as above.
\item  $E2Z_{SV}$ : Back Composition  sentence vector similarity.  A round trip translation is done for each n-best hypothesis to translate it back to Chinese. Then we use sentence vector similarity to measure the similarity between the original source and the recomposed source.
\end{itemize}

\section{Experiments}
\label{sec:experiments}

In this section, we first introduce the data and experimental setup used in our experiments, and then evaluate each of the systems introduced in Section~3, both independently and after system combination and re-ranking.

\subsection{Data and Experimental Setup}
We use all of the available parallel data for the WMT17 Chinese-English translation task. This consists of about 332K sentence pairs from the News Commentary corpus, 15.8M sentence pairs from the UN Parallel Corpus, and 9M sentence pairs from the CWMT Corpus. We further filter the bilingual corpus according to the following criteria:
\begin{itemize}
\item Both the source and target sentences should contain at least 3 words and at most 70 words.
\item Pairs where (source length $< 1.3*$target length or target length $< 1.3*$source length) are removed.
\item Sentences with illegal characters (such as URLs, characters of other languages) are removed. 
\item Chinese sentences without any Chinese characters are removed.
\item Duplicated sentence pairs are removed.
\end{itemize}

After filtration, we are left with 18M bilingual sentence pairs. We use the Chinese and English language models trained on the 18M sentences of bilingual data to filter the monolingual sentences from ``News Crawl: articles from 2016'' and ``Common Crawl'' provided by WMT17 using CED \cite{Moorelewis}. After filtering, we retain about 7M English and Chinese monolingual sentences. The monolingual data will be deployed in both dual learning and back-translation setups through the experiments.

Newsdev2017 is used as the development set and Newstest2017 as the test set. 
All the data (parallel and monolingual) have been tokenized and segmented into subword symbols using byte-pair encoding (BPE)~\cite{sennrich2015neural}.
The Chinese data has been tokenized using the Jieba tokenizer\footnote{https://github.com/fxsjy/jieba}. English sentences are tokenized using the scripts provided in Moses. We learn a BPE model  with 32K merge operations, in which 44K and 33K sub-word tokens are adopted as source and target vocabularies separately. 
 
%


\subsection{Experimental Results}

The Transformer model \cite{Vaswani2017AttentionIA} is adopted as our baseline.
Unless otherwise mentioned, all translation experiments use the following hyper-parameter settings based on Tensor2Tensor Transformer-big settings v1.3.0\footnote{https://github.com/tensorflow/tensor2tensor/blob/master/tensor2tensor/models/transformer.py}. This corresponds to a 6-layer transformer with a model size of 1024, a feed forward network size ($d_{ff}$) of 4096, and 16 heads. All models are trained on 8 Tesla M40 GPUs for a total of 200K steps using the Adam~\cite{Kingma2014AdamAM} algorithm. The initial learning rate is set to 0.3 and decayed according to the ``noam'' schedule as described in \cite{Vaswani2017AttentionIA}.During training, the batch size is set to 5120 words per batch and checkpoints are created every 60 minutes. All  results are reported on averaged parameters of the last 20 checkpoints. At test time, we use a beam of 8 and a length penalty of 1.0. All reported scores are computed using sacreBLEU v1.2.3,\footnote{https://github.com/awslabs/sockeye/tree/master/contrib/sacrebleu} which calculates tokenization-independent BLEU \cite{papineni2002bleu}.\footnote{sacreBLEU signature: \texttt{BLEU+case.mixed+lang.zh-en+numrefs.1+smooth.exp\_+test.wmt17/improved+tok.13a+version.1.2.3}}

The first section of Table \ref{exp_results} shows the results for the baselines. First we compare with the Sogou system \cite{wang-EtAl:2017:WMT}, which was the best result reported at WMT~2017 evaluation campaign. Though Sogou is an ensemble of many systems, we reference it here for comparison. The rest of the systems reported in the table are single systems. Our baseline system, labeled \textbf{Base}, is  trained on 18M sentences. \textbf{BT} is adding the back-translated data to the baseline.

\begin{table}[ht]
\centering

\begin{tabular}{@{}llr@{}}
\textbf{SystemID} &\textbf{Settings}                                       & \textbf{BLEU} \\
\toprule
Sogou &WMT~2017 best result {\cite{wang-EtAl:2017:WMT}}    & 26.40    \\
Base & Transformer Baseline    & 24.2       \\ 
BT & +Back Translation     & 25.57                \\
\midrule
DL & BT + Dual Learning     & 26.51 \\
DLDN & BT + Dual Learning + Deliberation Nets & 27.40\\
DLDN2 & DLDN without first decoder reranking & 27.20\\
DLDN3 & BT+ Dual Learning + R2L sampling & 26.88\\
DLDN4 & BT+ Dual Learning + Bi-NMT  & 27.16\\
\midrule
AR & BT + Agreement Regularization   & 26.91          \\ 
ARJT & BT + Agreement Regularization + Joint Training  & 27.38\\
ARJT2 & ARJT + dropout=0.1 & 27.19\\
ARJT3 & ARJT + dropout=0.05 & 27.07\\
ARJT4 & ARJT + dropout=0.01 & 26.98\\
\bottomrule

\end{tabular}

\caption{Automatic (BLEU) evaluation results on the WMT~2017 Chinese-English test set}
\label{exp_results}
\end{table}


\paragraph{Experimental Results of Dual Learning and Deliberation Networks}\mbox{}\\

Our Dual Learning system consists of a Zh$\to$En model and an En$\to$Zh model, each adopting the same model configuration as the baseline (Base). For the deliberation network, the encoder and the first-pass decoder are initialized from the Zh$\to$En model in the Dual Learning system, and the second pass decoder share the same model structures with the first-pass decoder. 
The evaluation results of the Dual Learning and Deliberation Network systems on WMT~2017 Chinese-English test set are listed in the second section of Table~\ref{exp_results}. Dual Learning makes more efficient use of the monolingual sentences and exploits the duality between Zh$\to$En and En$\to$Zh translation directions. Based on system \textbf{BT}, the Dual Learning system \textbf{DL} achieves 26.51 BLEU, a 0.94 point improvement over the \textbf{BT} system, and outperforms the best ensemble result of 26.40 in the WMT~2017 Chinese-English challenge . The Deliberation Network is further applied to the Dual Learning system, which is denoted as \textbf{DLDN}. The Deliberation Network aims to improve sentence generation quality by incorporating the global information provided by a first pass decoder. The \textbf{DLDN} system further achieves a BLEU score of 27.40, a 0.89 BLEU score improvement over the already strong \textbf{DL} system.

We also explore some variants of our \textbf{DL} and \textbf{DLDN} systems, denoted as \textbf{DLDN2/3/4} in the second section of Table~\ref{exp_results}. 
In \textbf{DLDN}, we use both the first and second pass decoders to rerank the generated sentence and choose the top-1 result. In system \textbf{DLDN2}, we then remove this reranking to see how the performance changes, yielding a 27.20 BLEU score, a 0.2 point drop. In system \textbf{DLDN3}, we replace the Deliberation Network with R2L sampling. R2L sampling is a data augmentation technique where we first train a Zh$\to$En model that generates sentences in a right-to-left(R2L) manner by reversing the target sentence in the training data, and use the R2L model to sample English sentences given monolingual Chinese sentences. We can see that adding R2L sampling to Dual Learning indeed brings BLEU score improvements, but performs worse than the Deliberation Network. In system \textbf{DLDN4}, we further add Bi-NMT, which bidirectionally generates candidate sentences in a single model, on the \textbf{DL} system and achieve 27.16 BLEU score.

\paragraph{Experimental Results of Agreement Regularization and Joint Training}\mbox{}\\

Data enhancement has been shown to improve NMT performance. We proposed the agreement regularization approach to explore data enhancement by using a right to left model to encourage consensus translations. The existing back-translation method is also one of the data enhancement approaches that leverages monolingual target data to generate synthetic bilingual data. Extending the back-translation approach, our proposed joint-training approach interactively makes data enhancement by boosting source-to-target and target-to-source NMT systems. Eventually, the unified joint training framework, denoted as \textbf{ARJT}, is used to integrate the agreement regularization approach, the back translation approach, and the joint training approach to further improve the performance of NMT systems.
The evaluation results of the agreement regularization and the unified joint training are listed in the third section of Table \ref{exp_results}. Compared to \textbf{BT}, our agreement regularization can achieve improvements of 1.34  BLEU points. Adding the joint training can bring this up to a 1.81 gain. 

We also explore several variants of our \textbf{ARJT}  system, denoted as \textbf{ARJT2/3/4} in Table \ref{exp_results}. We vary the dropout probability in order to explore the interaction between dropout regularization and agreement regularization. Unlike \textbf{ARJT}, these variants don't use the validation set for early stopping. 

\paragraph{Experimental Results of Data Selection}\mbox{}\\

In addition to our results using the WMT training data, we also explore training our system on a larger corpus. We experimented with  100M parallel sentences drawn from UN data, Open Subtitles and Web crawled data. 
It is worth noting that the  experiments reported  in Table \ref{exp_results}  were constrained data experiments limited to WMT17 official data only. While the experiments reported in Table \ref{exp_dataselect} are unconstrained systems using additional data.

First we apply word alignment heuristics to filter very noisy data.  This filters out around 10\%  of the data. Then we apply Cross-Entropy data selection \cite{Moorelewis} and \cite{Axelrod} to order the sentences based on their relevance to the CWMT part of the WMT data. We then select a specific number of sentences pairs by rank. 

In a separate experiment, we also apply the SentVec similarity filtering, described in Section \ref{sec.ds},  to select the same amount of data and measure its effect. We use  a cutoff threshold of the cosine similarity of 0.2. We train the unified bi-lingual encoder on a selected subset of the data that is known to be of good quality and in the relevant domain, specifically, the CWMT data of 9M sentence pairs. Since the system is trained to translate in both directions, it is effectively trained on on 18M sentence pairs.

Table \ref{exp_dataselect} shows the results of data selection. \textbf{Base8K} is using baseline data and back translated data, however it uses a larger model architecture that we found to work better with larger data sets. \textbf{Base8K} uses  6-layer transformer with a model size of 1024, a Feed Forward Network size ($d_{ff}$) of 8192, and 16 heads. All models reported in Table \ref{exp_dataselect}  are trained for  $300K$ steps with minibatch of $3500$ on 8 GPUs. We average the last $20$ checkpoints as before and decode with beam size of $8$ and length penalty of $1.0$ similar to the setup above.

\textbf{CED1} and \textbf{CED2} add 35M sentences and 50M sentences respectively to \textbf{Base8k}. \textbf{SV1}
and \textbf{SV2} added the same amount of data selected by SentVec similarity discussed in Section \ref{sec.ds}.  \textbf{SV3} and  \textbf{SV4} experimented with varying the dropout ratio to measure its impact with the larger training data and model architecture. Generally the systems using  SentVec similarity filtering  achieve improvements up to 1.5  BLEU points over \textbf{Base8K} and nearly 1 BLEU point as compared to systems using the same amount of CED-selected data. We conclude that SentVec similarity filtering is a helpful approach since it filters out noisy data which is hard to identify. Since SentVec prevents data with partial and low-quality translation from negatively impacting the system. Furthermore, the proposed approach helps select relevant data similar to CWMT data.

\begin{table}[ht]
\centering
\begin{tabular}{@{}llr@{}}
\textbf{SystemID} &\textbf{Settings}                                       & \textbf{BLEU} \\
\toprule
Base & Transformer Baseline    & 24.2       \\ 
BT & +Back Translation     & 25.57                \\
Base8K & BT + 8K $d_{ff}$ & 26.13 \\
 
\midrule
CED1 & Base8K + 35M CED + dropout=0.1   &  26.68 \\
CED2 & Base8K + 50M CED + dropout=0.1   & 26.61  \\
\midrule 
SV1 & Base8K + 35M + dropout=0.1 &  27.60\\
SV2 & Base8K + 50M + dropout=0.1 & 27.45\\
SV3 & Base8K + 35M + dropout=0.2 & 27.67\\
SV4 & Base8K + 50M + dropout=0.2 & 27.49\\
\bottomrule

\end{tabular}
\caption{Evaluation Data selection results on the WMT~2017 Chinese-English test set}
\label{exp_dataselect}
\end{table}

\paragraph{Experimental Results of Systems Combination}\mbox{}\\

We experiment with system combination of n-best lists generated from various systems discussed above with  8 hypothesis from each system. We use various features to re-rank the systems hypothesis as described in Section \ref{sec.sc}. 
As shown in Table \ref{exp_systemcomb}, combining the set of  heterogeneous systems are complementary and achieved the highest results. We have experimented with many configurations and features for systems combination, we found out that the most helpful scoring features are: $SYS_{Score}$, $LM_{Score}$, $R2L_{score}$,  $R2L_{SV}$  and $E2Z_{SV}$. This is  quite surprising  since the combined systems were focusing on modeling similar features. This may be due to the fact that the models are learning complimentary features, so they have extra capacity for complementing each other.

We think it would be useful to combine all proposed approaches in a single system. However, we leave this as a future work item.

\begin{table}[ht]
\centering
\begin{tabular}{@{}rll@{}}
\textbf{SystemID} &\textbf{Settings}                                       & \textbf{BLEU} \\
\toprule
Combo-1 & SV1, SV2, SV3     & 27.84       \\ 
Combo-2 & DLDN2, DLDN3, DLDN4 & 27.92 \\
Combo-3 &  ARJT2, ARJT3, ARJT4 + 3 identical systems with different initialization    & 27.82 \\
\midrule
Combo-4 & SV1, SV2, SV3, ARJT1, ARJT2, ARJT3, DLDN2, DLDN3, DLDN4 & 28.46\\
Combo-5 & SV1, SV2, SV3, ARJT2, DLDN2, DLDN4 & 28.32 \\
Combo-6 & SV1, SV2, SV4, ARJT2, ARJT3, ARJT4, DLDN2, DLDN3, DLDN4     & 28.42 \\
\bottomrule
\end{tabular}
\caption{System combination results on the WMT~2017 Chinese-English test set}
\label{exp_systemcomb}
\end{table}


\newcommand{\ExptNine}{Combo-1}
\newcommand{\ExptNLC}{Combo-2}
\newcommand{\Single}{SV1}

\newcommand{\ComboA}{\textsc{Combo-4}}
\newcommand{\ComboB}{\textsc{Combo-5}}
\newcommand{\ComboC}{\textsc{Combo-6}}

\newcommand{\Sogou}{\textsc{Sogou}}
\newcommand{\Microsoft}{\textsc{Online-A-1710}}
\newcommand{\Google}{\textsc{Online-B-1710}}

\newcommand{\MetaA}{\textsc{Meta-1}}
\newcommand{\MetaB}{\textsc{Meta-2}}
\newcommand{\MetaC}{\textsc{Meta-3}}

\newcommand{\SubsetA}{\textsc{Subset-0}}
\newcommand{\SubsetB}{\textsc{Subset-1}}
\newcommand{\SubsetC}{\textsc{Subset-2}}
\newcommand{\SubsetD}{\textsc{Subset-3}}
\newcommand{\SubsetE}{\textsc{Subset-4}}

\newcommand{\newstest}{\textsc{newstest2017}}
\newcommand{\RefWMT}{\textsc{Reference-WMT}}
\newcommand{\RefHT}{\textsc{Reference-HT}}
\newcommand{\RefPE}{\textsc{Reference-PE}}

\section{Human Evaluation Results}
\label{sec:human-eval-results}

\nocite{edgington1980validity}
\nocite{wilcoxon1945individual}

Table~\ref{evalresults} presents the results from our large scale human evaluation campaign. Based on these results we claim that we have achieved human parity according to Definition~\ref{def:human-parity}, as our research systems are indistinguishable from human translations.

In the table, systems in higher clusters significantly outperform all systems in lower clusters according to Wilcoxon rank sum test at p-level $p \leq 0.05$, following WMT17. Systems in the same cluster are ordered by $z$ score---which is defined as the signed number of standard deviations an observation is above the mean, computed on the annotator level to address different annotation behavior---but considered tied w.r.t. quality.

\begin{table}
\centering

\begin{tabular}{@{}rrrl@{}}
$\#$ & Ave $\%$ & Ave $z$ & System \\
\toprule
1 & 69.0 & 0.237 & \ComboC \\
  & 68.5 & 0.220 & \RefHT \\
  & 68.9 & 0.216 & \ComboB \\
  & 68.6 & 0.211 & \ComboA \\
\midrule
2 & 67.3 & 0.141 & \RefPE \\
\midrule
3 & 62.3 & -0.094 & \Sogou \\
  & 62.1 & -0.115 & \RefWMT \\
\midrule
4 & 56.0 & -0.398 & \Microsoft \\
  & 54.1 & -0.468 & \Google \\
\bottomrule
\end{tabular}

\caption{\textbf{Human Evaluation Results} for at least $n \geq 1,827$ assessments per system show that our research systems \ComboA{}, \ComboB{}, and \ComboC{} achieve human parity according to definition~\ref{def:human-parity} as they are not distinguishable from \RefHT{}, which is a human translation. All our research systems significantly outperform \RefPE{}, which is based on human post-editing of machine translation output, and the original \RefWMT{}, which is again a human translation. \emph{\#} denotes the ranking cluster, \emph{Ave $\%$} the averaged raw score $r \in [0, 100]$, and \emph{Ave $z$} the standardized $z$ score. $n \geq x$ denotes that we collected at least $x$ assessments per system for the respective evaluation campaign. This is referred to as \MetaA{} in Table~\ref{meta1}.}
\label{evalresults}
\end{table}

\subsection{Human Evaluation Setup}

As discussed in Section~\ref{sec:human-parity} our evaluation methodology is based on \emph{source-based direct assessment} as described in \cite{IWSLT17}. We use an updated version of Appraise \cite{Appraise}, the same tool which is used in the human evaluation campaign for the Conference on Machine Translation (WMT).\footnote{This version of Appraise will also be used to run the WMT18 evaluation campaigns. Source code will be released to the public in time for WMT18, as in previous years.} See \cite{bojar-EtAl:2017:WMT1} for more details on last year's WMT17 results and evaluation.

\bigskip\noindent The main differences to the WMT17 campaign are:
\begin{enumerate}
\item Our evaluation is based on quality assessment of translations with respect to the source text, not a reference translation. To do this, we hire bilingual crowd workers;


\item We enforce full system coverage for the evaluation samples. This means that for every segment we get human scores for all systems under investigation;

\item We require redundancy so that for every annotation task (also referred to as ``HIT'' in other direct assessment publications) we collect scores from three annotators.
\end{enumerate}

The latter two changes have been introduced to strengthen our results, by adding additional redundancy. Direct assessment as an estimator of general system quality does not require these, but in the context of achieving human parity, extra layers of fully comparable segment scores enable more thorough external validation. We intend to release all data related to the final human parity evaluation campaigns, so this data will become available for independent inspection by the research community.

\subsection{Benchmark Translations}
\label{comparesystems}
We compare our research systems against the following sets of translations. These sets have been kept stable across all evaluation campaigns, allowing us to track research results over time.

\begin{description}
\item[Reference-HT] vendor-created human translations of \emph{newstest2017}. Translators were instructed to translate from scratch, i.e., without using any online translation engines;\footnote{Of course, there are sentences for which the human translation matches Google Translate or Microsoft Translator machine translation output. Relative to the overlap for the post-editing-based reference, this is negligible.}

\item[Reference-PE] vendor-created human post-editing output, based on Google Translate machine translation results;

\item[Reference-WMT] Original \emph{newstest2017} reference released after WMT17. The original WMT17 reference translation for \emph{newstest2017} is known to contain errors, so we decided to add it to the set of evaluated systems. This allows us to get external validation for the quality of our two human references;

\item[Online-A-1710] Microsoft Translator production system, collected on October 16, 2017;

\item[Online-B-1710] Google Translate production system, collected on October 16, 2017;

\item[Sogou] The Sogou Knowing NMT system, which performed best at last year's WMT17 Conference on Machine Translation (WMT) shared task on news translation \cite{Sogou}.
\end{description}

Note that the benchmark human references were not available to the system developers.
Also, the presented set of translation systems affects human-perceived quality (both based on the total number and distribution of quality across systems), so we do not expect scores to be comparable across campaigns. The question of comparability of raw direct assessment scores over time is an open research problem still, so we take a conservative approach and do not compare them. Scores within a single campaign are reliable. We also assume that standardized scores for the same set of translation systems should be fairly comparable.

\subsection{Guarding Against Confounds}
\label{confounds}

Whenever trying to draw a conclusion based on a pair of different translations, we must avoid measuring the effects of extraneous variables that can confound the experimental variables we wish to measure~\cite{clark2011better}. For example, when comparing the translation quality by varying how it is produced (human translation versus automatic translation), we do not wish our measurements of translation quality to be influenced by external factors, e.g., perhaps a human translator did a poor job when translating a few sentences or an automatic translation system happens to be exceptionally good at translating a particular subset of sentences.

\bigskip \noindent In this work, we specifically control for the effects of several potential extraneous variables:
\begin{itemize}

\item \textbf{Variability of quality measure} \emph{How sensitive is our quality measure (direct assessment) to different subsets of the data?} We answer this by running redundant evaluation campaigns across different subsets of the data.

\item \textbf{Test set selection} \emph{Would we likely obtain the same result on slightly different test data?} We control for this by running redundant large-scale human evaluation campaigns under several configurations to replicate results (Section~\ref{campaigns}).

\item \textbf{Annotator errors} \emph{What if some annotators become inattentive, unfairly improving or damaging the score of one system over the other?} To control for this effect, we use rejection sampling when gathering human assessments by occasionally showing annotators examples where one sentence is intentionally and noticeably worse; annotators that fail to detect these are excluded from the data, ensuring that human judgments are high quality.

\item \textbf{Annotator inconsistency} \emph{What if the annotators produce different scores given the same data? Would using different annotators still lead to the same conclusion?} To control for this, our evaluation campaigns directly include multiple evaluators.

\item \textbf{Choice of systems} \emph{Was this particular system combination somehow ``lucky'', or would similar combinations also lead to the same conclusion?} To answer this question, we include multiple system combinations with varying sets of input systems. (Section~\ref{campaigns})
\end{itemize}

\subsection{Evaluation Campaigns}
\label{campaigns}

\bigskip\noindent We conduct the following evaluations:
\begin{description}

\item[Annotator variability study] To measure this, we repeat the same evaluation campaign three times. All data is collected on the same subset. We allow annotator overlap but do not enforce it. In the end, we had a near complete annotator overlap, likely due to the timing of our campaigns.\footnote{To complete so many campaigns in such a short time, it was easier to attract crowd workers when they knew they could earn more by completing several campaigns. Combined with our reliability testing, this motivation likely had a positive impact on annotation fidelity and quality.}  We refer to this as \textsc{Eval Round 1}, on evaluation sample \SubsetB;

\item[Data variability study] Our data subsets are randomly selected from the source data. Still, the actual subset could affect results in our favor. To counter this, we conduct three additional evaluation campaigns on three completely different subsets of data.  We refer to this as \textsc{Eval Round 2}, on evaluation samples \SubsetC,  \SubsetD, and  \SubsetE.
\end{description}

As the set of systems for all these campaigns does not change, results are theoretically comparable, so we can also report synthesized, joint scores, for both dimensions in isolation and in combined form.

\begin{samepage}
Evaluation campaign parameters are as follows:
\begin{itemize}[noitemsep]
\item Annotators: 15
\item Tasks: 20
\item Redundancy: 3
\item Tasks per annotator: 4 (about 2 hours of work)
\item Systems: 9
\item Data points: 4,200 (at least\footnote{Note that as we annotate on unique translation output only, there is a chance that more data points are collected.} 466 per system)
\end{itemize}
\end{samepage}

The set of systems for the final evaluation campaigns consists of the following systems:
\begin{itemize}[noitemsep]
\item References: \RefHT, \RefPE, \RefWMT
\item Production: \Microsoft, \Google
\item WMT17: \Sogou
\item Candidates: \ComboA, \ComboB, \ComboC
\end{itemize}

After completion of all six evaluation campaigns, we have collected at least 25,200 data points (i.e., segment scores) or at least 2,520 per system. This is comparable to the amount of annotations collected for last year's WMT17 evaluation campaign (2,421 assessments per system). We report results for individual campaigns and our final \emph{synthesized}, joint meta-campaign:

\begin{description}[noitemsep]
\item[\MetaA] We combine assessments from evaluation campaigns \textsc{Eval Round 1}a--c, on evaluation sample SubsetB, effectively increasing data points by a factor of 3x. Note that this is fair as result clusters are based on standardized scores which can fairly be computed if all annotators are exposed to exactly the same segments per system.


\end{description}

While it is also possible to combine data across subsets, we choose not to do this as this potentially affects standardization of annotator scores. For \MetaA, due to the identical assignment of annotators to segments, we have a guarantee that standardization is reliable.



\subsection{Annotator Variability Results}
\begin{description}
\item[\SubsetB, first iteration] Table~\ref{eval2a} shows the results of our first evaluation round on \SubsetB. Note how our research systems outperform \Sogou~ and both \RefWMT~ and \RefPE. Based on this clustering it becomes clear that there must be quality issues with the original \RefWMT~ reference. All three systems \ComboA, \ComboB, and \ComboC~ achieve human parity with \RefHT. We collected at least $n \geq 609$ assessments per system.

\item[\SubsetB, second iteration] Table~\ref{eval2b} shows the results for our second evaluation round on \SubsetB. This time, annotators do not see a significant difference between our research systems and \RefPE. Consequently, \RefHT and all three systems \ComboA, \ComboB, and \ComboC~ end up in the same cluster as \RefPE. All these systems outperform \Sogou~ and \RefWMT. As in the previous round, online systems \Microsoft~ and \Google~ perform worst.

\item[\SubsetB, third iteration] Table~\ref{eval2c} shows the results for our third evaluation round on \SubsetB. Similar to the second round, we do not observe a significant difference between \RefPE~ and our research systems. Again, \RefHT, all three systems \ComboA, \ComboB, and \ComboC, and \RefPE{} end up in the top cluster. \Sogou~ and \RefWMT~ end in the third cluster, outperforming \Microsoft~ and \Google. Again, the latter are not significantly different w.r.t human perceived quality.
\end{description}

\subsection{Data Variability Results}
\begin{description}
\item[\SubsetC] Table~\ref{eval3a} shows the results for our evaluation on \SubsetC. Annotators seem to have a preference for \RefHT~ over \ComboA, \ComboB, and \ComboC, but not significantly so. All four systems outperform \RefPE, which itself outperforms all other systems. \Sogou~ ends up in its own cluster, significantly better than \RefWMT~ and the two online systems \Microsoft~ and \Google. We collected at least $n \geq 607$ assessments per system.

\item[\SubsetD] Table~\ref{eval3b} shows the results for our evaluation on \SubsetD. This one is interesting as it is the only evaluation round which shows \RefPE~ on top, based on its $z$ score. Otherwise, we continue to see \RefHT, \ComboA, \ComboB, and \ComboC~ in the top cluster. \Sogou~ and \RefWMT~ are indistinguishable for this subset and both outperform the two online systems, \Microsoft~ and \Google. We collected at least $n \geq 610$ assessments per system. 

\item[\SubsetE] Table~\ref{eval3c} shows the results for our evaluation on \SubsetE. Again, our research systems \ComboA, \ComboB, and \ComboC~ are indistinguishable from \RefHT{} and \RefPE. There is no significant difference in quality between these five systems. \Sogou{} and \RefWMT~ outperform the online systems \Microsoft{} and \Google. We collected at least $n \geq 649$ assessments per system. 
\end{description}


\newgeometry{left=1cm,right=1cm,bottom=2cm,top=1cm}

\begin{table}
\centering

\begin{subtable}[b]{.3\textwidth}
\centering
\footnotesize

\begin{tabular}{@{}rrrl@{}}
$\#$ & Ave $\%$ & Ave $z$ & System \\
\toprule
1 & 69.9 & 0.256 & \ComboC \\
  & 69.8 & 0.233 & \ComboA \\
  & 69.9 & 0.230 & \ComboB \\
  & 68.6 & 0.186 & \RefHT \\
  & 67.6 & 0.129 & \RefPE \\
\midrule
2 & 63.3 & -0.095 & \Sogou \\
  & 62.1 & -0.132 & \RefWMT \\
\midrule
3 & 57.0 & -0.383 & \Microsoft \\
  & 54.1 & -0.494 & \Google \\
\bottomrule
\end{tabular}

\caption{\SubsetB, $n \geq 609$}
\label{eval2a}
\end{subtable}
~
\begin{subtable}[b]{.3\textwidth}
\centering
\footnotesize

\begin{tabular}{@{}rrrl@{}}
$\#$ & Ave $\%$ & Ave $z$ & System \\
\toprule
1 & 68.6 & 0.233 & \RefHT \\
  & 68.6 & 0.225 & \ComboC \\
  & 68.6 & 0.217 & \ComboB \\
  & 68.3 & 0.207 & \ComboA \\
  & 67.4 & 0.154 & \RefPE \\
\midrule
2 & 61.9 & -0.105 & \Sogou \\
  & 62.1 & -0.113 & \RefWMT \\
\midrule
3 & 55.7 & -0.399 & \Microsoft \\
  & 53.9 & -0.468 & \Google \\
\bottomrule
\end{tabular}

\caption{\SubsetB, second iteration}
\label{eval2b}
\end{subtable}
~
\begin{subtable}[b]{.3\textwidth}
\centering
\footnotesize

\begin{tabular}{@{}rrrl@{}}
$\#$ & Ave $\%$ & Ave $z$ & System \\
\toprule
1 & 68.5 & 0.240 & \RefHT \\
  & 68.4 & 0.229 & \ComboC \\
  & 68.1 & 0.201 & \ComboB \\
  & 67.7 & 0.194 & \ComboA \\
  & 66.8 & 0.141 & \RefPE \\
\midrule
2 & 61.8 & -0.083 & \Sogou \\
  & 62.0 & -0.100 & \RefWMT \\
\midrule
3 & 55.2 & -0.413 & \Microsoft \\
  & 54.3 & -0.442 & \Google \\
\bottomrule
\end{tabular}

\caption{\SubsetB, third iteration}
\label{eval2c}
\end{subtable}
\\[1cm]
\begin{subtable}[b]{.3\textwidth}
\centering
\footnotesize

\begin{tabular}{@{}rrrl@{}}
$\#$ & Ave $\%$ & Ave $z$ & System \\
\toprule
1 & 68.6 & 0.212 & \RefHT \\
  & 68.2 & 0.200 & \ComboB \\
  & 67.9 & 0.182 & \ComboA \\
  & 67.9 & 0.177 & \ComboC \\
\midrule
2 & 64.8 & 0.044 & \RefPE \\
  & 62.5 & -0.061 & \Sogou \\
\midrule
3 & 59.6 & -0.200 & \RefWMT \\
  & 58.4 & -0.277 & \Microsoft \\
  & 55.7 & -0.353 & \Google \\
\bottomrule
\end{tabular}

\caption{\SubsetC, $n \geq 607$}
\label{eval3a}
\end{subtable}
~
\begin{subtable}[b]{.3\textwidth}
\centering
\footnotesize

\begin{tabular}{@{}rrrl@{}}
$\#$ & Ave $\%$ & Ave $z$ & System \\
\toprule
1 & 67.4 & 0.251 & \RefHT \\
  & 67.1 & 0.247 & \RefPE \\
  & 65.3 & 0.147 & \ComboC \\
  & 64.9 & 0.106 & \ComboA \\
  & 64.3 & 0.091 & \ComboB \\
\midrule
2 & 61.1 & -0.065 & \Sogou \\
  & 59.6 & -0.119 & \RefWMT \\
\midrule
3 & 55.3 & -0.351 & \Microsoft \\
  & 54.4 & -0.377 & \Google \\
\bottomrule
\end{tabular}

\caption{\SubsetD, $n \geq 650$}
\label{eval3b}
\end{subtable}
~
\begin{subtable}[b]{.3\textwidth}
\centering
\footnotesize

\begin{tabular}{@{}rrrl@{}}
$\#$ & Ave $\%$ & Ave $z$ & System \\
\toprule
1 & 66.6 & 0.254 & \RefHT \\
  & 65.2 & 0.179 & \ComboC \\
  & 64.4 & 0.151 & \ComboB \\
  & 64.2 & 0.147 & \ComboA \\
  & 63.4 & 0.127 & \RefPE \\
\midrule
2 & 60.5 & -0.030 & \Sogou \\
  & 60.1 & -0.074 & \RefWMT \\
\midrule
3 & 53.4 & -0.367 & \Microsoft \\
  & 51.7 & -0.455 & \Google \\
\bottomrule
\end{tabular}

\caption{\SubsetE, $n \geq 649$}
\label{eval3c}
\end{subtable}
\\[1cm]
\begin{subtable}[b]{.3\textwidth}
\centering
\footnotesize

\begin{tabular}{@{}rrrl@{}}
$\#$ & Ave $\%$ & Ave $z$ & System \\
\toprule
1 & 69.0 & 0.237 & \ComboC \\
  & 68.5 & 0.220 & \RefHT \\
  & 68.9 & 0.216 & \ComboB \\
  & 68.6 & 0.211 & \ComboA \\
\midrule
2 & 67.3 & 0.141 & \RefPE \\
\midrule
3 & 62.3 & -0.094 & \Sogou \\
  & 62.1 & -0.115 & \RefWMT \\
\midrule
4 & 56.0 & -0.398 & \Microsoft \\
  & 54.1 & -0.468 & \Google \\
\bottomrule
\end{tabular}

\caption{\MetaA, $n \geq 1,827$}
\label{meta1}
\end{subtable}
\\[1cm]




\caption{Complete results for our three iterations over \SubsetB~(\ref{eval2a}, \ref{eval2b}, \ref{eval2c}) and our evaluation campaigns for \SubsetC~(\ref{eval3a}), \SubsetD~(\ref{eval3b}), and \SubsetE~(\ref{eval3c}). We also show results for combined data for \MetaA~(\ref{meta1}) combining annotations from all iterations over \SubsetB. \emph{\#} denotes the ranking cluster, \emph{Ave $\%$} the averaged raw score $r \in [0, 100]$, and \emph{Ave $z$} the standardized $z$ score. $n \geq x$ denotes that we collected at least $x$ assessments per system for the respective evaluation campaign. All campaigns involved $a=15$ annotators. Systems in higher clusters significantly outperform all systems in lower clusters according to Wilcoxon rank sum test at p-level $p \leq 0.05$, following WMT17. Systems in the same cluster are ordered by $z$ score but considered tied w.r.t. quality.}
\label{babeleval}
\end{table}

\restoregeometry

\FloatBarrier

\begin{table}[h]
\centering
\small
\begin{tabular}{@{}lrrrcrrrcr@{}}
\toprule
\multirow{2}{*}[-2pt]{System} & \multicolumn{3}{c}{refs=1} && \multicolumn{3}{c}{refs=2} && \multicolumn{1}{c}{refs=3} \\
\cmidrule(l){2-4} \cmidrule(l){6-8} \cmidrule(l){10-10}
  & WMT & PE & HT && WMT+PE & WMT+HT & PE+HT && WMT+PE+HT \\
\midrule
\Microsoft & 24.38 & 28.82 & 17.12 && 36.53 & 32.17 & 35.33 && 41.21 \\
\Google    & 33.56 & 46.97 & 17.70 && 56.45 & 40.55 & 51.78 && 59.37 \\
\Sogou     & 26.37 & 30.69 & 19.71 && 38.67 & 35.47 & 38.19 && 44.18 \\
\midrule
\ComboA    & 28.30 & 29.79 & 20.47 && 39.53 & 37.73 & 38.43 && 45.62 \\
\ComboB    & 28.18 & 29.61 & 20.48 && 39.32 & 37.54 & 38.15 && 45.32 \\
\ComboC    & 28.07 & 29.90 & 20.70 && 39.39 & 37.77 & 38.45 && 45.64 \\
\bottomrule
\end{tabular}

\caption{BLEU scores against single or multiple references. WMT is \RefWMT, PE is \RefPE, HT is \RefHT. Scoring based on sacreBLEU v1.2.3, with signature \texttt{BLEU+case.mixed+numrefs.1+smooth.exp+tok.13a+version.1.2.3} for refs=1. Signature changes to \texttt{numrefs.2} and \texttt{numrefs.3} for refs=2 and refs=3, respectively. Note how different scores for \RefWMT~ and \RefPE~ are compared to \RefHT~ and how these compare to our findings reported in Table~\ref{babeleval}. This emphasizes the need for human evaluation.}
\label{babelbleu}
\end{table}

\subsection{Data Release}
\label{datarelease}
We have released\footnote{All Translator human parity data is available here: \url{http://aka.ms/Translator-HumanParityData}} all data from the human evaluation campaigns to 1) allow external validation of our claim of having achieved human parity and 2) to foster future research by releasing two additional human references for the \RefWMT~ test set.

The release package contains the following items:

\begin{description}
\item[New references for \newstest] Two new references for \newstest, one based on human translation from scratch (\RefHT), the other based on human post-editing (\RefPE). Table~\ref{babelbleu} reports the BLEU scores for single and multi reference use with sacreBLEU;

\item[Human parity translations] Output generated by our research systems \ComboA, \ComboB, and \ComboC;

\item[Online translations] Output from online machine translation service \Microsoft, collected on October 16, 2017;

\item[Human evaluation data] All data points collected in our human evaluation campaigns. This includes annotations for \SubsetB, \SubsetC, \SubsetD, and \SubsetE. We share the (anonymized) annotator IDs, segment IDs, system IDs, type ID (either \texttt{TGT} or \texttt{CHK}, the second being a repeated judgment for the first), raw scores $r \in [0, 100]$, as well as annotation start and end times.
\end{description}

We do not redistribute the following items:

\begin{description}
\item[\RefWMT~ test data] This is publicly available from the WMT17 website\footnote{\url{http://data.statmt.org/wmt17/translation-task/test.tgz}}. In this work, we used the source \texttt{newstest2017-zhen-src.zh} and the reference (as \RefWMT) \texttt{newstest2017-zhen-ref.en};

\item[\Sogou~ translation] This is publicly available from the WMT17 website as well\footnote{\url{http://data.statmt.org/wmt17/translation-task/wmt17-submitted-data-v1.0.tgz}}. We used \texttt{newstest2017.SogouKnowing-nmt.5171.zh-en} (as \Sogou).
\end{description}

The Appraise repository on GitHub\footnote{\url{https://github.com/cfedermann/Appraise}} contains code to recompute result clusters. We share this data in the hope that the research community might find it useful and also to ensure greatest possible transparency regarding the generation of the results presented in this paper.

\section{Human Analysis}
\label{human-analysis}
Lastly, a preliminary human error analysis was conducted over the output of the \ComboC{} system (the system that achieved the best results). We randomly sampled 500 sentences and annotated each translation with whether a specific error type was present. Following \cite{ErrorAnalysis}, we use 9 categories: Missing Words, Word Repetition, Named Entity, Word Order, Incorrect Words, Unknown Words, Collocation, Factoid, and Ungrammatical. The Named-Entity category is further subdivided into Person, Location, Organization, Event, and Other.

\begin{table}[htb]
\centering
\begin{tabular}{@{}p{4cm}l@{}}
\toprule
\textbf{Error Category}            & \textbf{Fraction [\%]}    \\
\midrule
Incorrect Words                    & 7.64                      \\
Ungrammatical                      & 6.33                      \\
Missing Words                      & 5.46                      \\
Named Entity                       & 4.38                      \\ 
\multicolumn{1}{r}{Person\hspace*{2em}}         & \multicolumn{1}{r}{1.53}  \\
\multicolumn{1}{r}{Location\hspace*{2em}}       & \multicolumn{1}{r}{1.53}  \\
\multicolumn{1}{r}{Organization\hspace*{2em}}   & \multicolumn{1}{r}{0.66}  \\
\multicolumn{1}{r}{Event\hspace*{2em}}          & \multicolumn{1}{r}{0.22}  \\
\multicolumn{1}{r}{Other\hspace*{2em}}          & \multicolumn{1}{r}{0.44}  \\
Word Order                         & 0.87                      \\
Factoid                            & 0.66                      \\
Word Repetition                    & 0.22                      \\
Collocation                        & 0.22                      \\
Unknown Words                      & 0                         \\
\bottomrule
\end{tabular}
\caption{Error distribution, as fraction of sentences that contain specific error categories.}
\label{my-ana}
\end{table}

Table \ref{my-ana} shows the distribution of the annotated errors as the fraction of sentences containing a specific error category. The four major error types are Missing words, Incorrect Words, Ungrammatical, and Named Entity. Each accounts for roughly 5\% of errors. This indicates that there is still room to improve machine translation quality via various approaches, such as modeling Missing Words \cite{tu2016modeling,AttentionFertility}, integration of high quality data for named-entity translation, as well as domain and topic adaptation for the issues of incorrect words and ungrammaticality.

\section{Discussion and Future Work}

In this paper, we described the techniques used in the latest Microsoft machine translation system to reach a new state-of-the-art. Our evaluation found that our system has reached parity with professional human translations on the WMT~2017 Chinese to English news task, and exceeds the quality of crowd-sourced references. 

We exploited the dual nature of the translation problem to better utilize parallel data as well as monolingual data in a more principled way. We utilized  joint training of source-to-target, and target-to-source systems to further improve on the duality of the translation task. We addressed the exposure bias problem in two ways: by two-pass decoding using Deliberation networks, as well as by agreement regularization and joint training of left-to-right, right-to-left systems. We trained a bilingual encoder to obtain bilingual sentence representations used to filter noisy data and select relevant data. We also found significant gains from combining multiple heterogeneous systems. 

We addressed the problem of defining and measuring the quality of human translations and near-human machine translations. We found that as translation quality has dramatically improved, automatic reference-based evaluation metrics have become increasingly problematic. We used direct human annotation to measure the quality of both human and machine translations.

We wish to acknowledge the tremendous progress in sequence-to-sequence modeling made by the entire research community that paved the road for this achievement. We have introduced a few new approaches that helped us to reach human parity for WMT2017 Chinese to English news translation task. At the same time, much work remains to be done, especially in domains and language-pairs that do not benefit from huge amounts of available data. 

\vfill

\bibliographystyle{acm}
\bibliography{sample}

\end{CJK*}
\end{document}